\documentclass[acmsmall]{acmart}

\usepackage[utf8]{inputenc} % allow utf-8 input
\usepackage[T1]{fontenc}    % use 8-bit T1 fonts
\usepackage{hyperref}       % hyperlinks
\usepackage{url}            % simple URL typesetting
\usepackage{booktabs}       % professional-quality tables
\usepackage{amsfonts}       % blackboard math symbols
\usepackage{nicefrac}       % compact symbols for 1/2, etc.
\usepackage{microtype}      % microtypography

\usepackage{setspace}

\usepackage{epsfig,bm,subfigure,graphicx,color}
\usepackage{subfigure}
\usepackage{wrapfig}

\newcommand{\boldI}{{\boldsymbol{I}}}

\newcommand{\boldW}{{\boldsymbol{W}}}
\newcommand{\boldX}{{\boldsymbol{X}}}

\newcommand{\bolda}{{\boldsymbol{a}}}
\newcommand{\boldb}{{\boldsymbol{b}}}

\newcommand{\bolde}{{\boldsymbol{e}}}

\newcommand{\boldh}{{\boldsymbol{h}}}

\newcommand{\boldv}{{\boldsymbol{v}}}

\newcommand{\boldx}{{\boldsymbol{x}}}

\newcommand{\boldz}{{\boldsymbol{z}}}

\newcommand{\boldtheta}{{\boldsymbol{\theta}}}

\usepackage{amsmath,amsfonts,amsthm,mathtools}
\usepackage{bbm}
\usepackage{lscape}

\usepackage{algorithm, algorithmic}

\newtheorem{theorem}{Theorem}
\newtheorem{lemma}[theorem]{Lemma}

\newtheorem{proposition}[theorem]{Proposition}

\theoremstyle{definition}

\newenvironment{prooftn}[1]{%
\renewcommand{\proofname}{Proof of Theorem {#1}}\proof}{\endproof}
\newenvironment{proofln}[1]{%
\renewcommand{\proofname}{Proof of Lemma {#1}}\proof}{\endproof}
\newenvironment{proofpn}[1]{%
\renewcommand{\proofname}{Proof of Proposition {#1}}\proof}{\endproof}
\newenvironment{sketch}{%
\renewcommand{\proofname}{Proof Sketch}\proof}{\endproof}

\usepackage{multirow}
\usepackage{tabularx}

\usepackage{pifont}
\newcommand{\cmark}{\ding{51}}%
\newcommand{\xmark}{\ding{55}}%

\AtBeginDocument{%
  \providecommand\BibTeX{{%
    \normalfont B\kern-0.5em{\scshape i\kern-0.25em b}\kern-0.8em\TeX}}}

\setcopyright{acmcopyright}
\acmJournal{TKDD}
\acmYear{2022} \acmVolume{16} \acmNumber{5} \acmArticle{92} \acmMonth{3} \acmPrice{15.00}\acmDOI{10.1145/3502733}

\begin{document}

\title{Constant Time Graph Neural Networks}

\author{Ryoma Sato}
\email{r.sato@ml.ist.i.kyoto-u.ac.jp}
\affiliation{%
  \institution{Kyoto University, RIKEN AIP}
  \country{Japan}
}

\author{Makoto Yamada}
\affiliation{%
  \institution{Kyoto University, RIKEN AIP}
  \country{Japan}
}

\author{Hisashi Kashima}
\affiliation{%
  \institution{Kyoto University, RIKEN AIP}
  \country{Japan}
}

\renewcommand{\shortauthors}{Sato et al.}

\begin{abstract}
The recent advancements in graph neural networks (GNNs) have led to state-of-the-art performances in various applications, including chemo-informatics, question-answering systems, and recommender systems. However, scaling up these methods to huge graphs, such as social networks and Web graphs, remains a challenge. In particular, the existing methods for accelerating GNNs either are not theoretically guaranteed in terms of the approximation error or incur at least a linear time computation cost. 
In this study, we reveal the query complexity of the uniform node sampling scheme for Message Passing Neural Networks, including GraphSAGE, graph attention networks (GATs), and graph convolutional networks (GCNs). Surprisingly, our analysis reveals that the complexity of the node sampling method is completely independent of the number of the nodes, edges, and neighbors of the input and depends only on the error tolerance and confidence probability while providing a theoretical guarantee for the approximation error. To the best of our knowledge, this is the first paper to provide a theoretical guarantee of approximation for GNNs within constant time. Through experiments with synthetic and real-world datasets, we investigated the speed and precision of the node sampling scheme and validated our theoretical results.
\end{abstract}




\maketitle

\section{Introduction}
\label{introduction}
Machine learning on graph structures has various applications, such as chemo-informatics \cite{MPNNs,DGCNN}, question answering systems \cite{RCGN,GENI}, and recommender systems \cite{PinSAGE,KGCN,KGAT,fan2019graph}. Recently, a novel machine learning model for graph data called graph neural networks (GNNs) ~\cite{gori, scarselli} demonstrated state-of-the-art performances in various graph learning tasks.
However, large scale graphs, such as social networks and Web graphs, contain billions of nodes, and even a linear time computation cost per iteration is prohibitive. 
Therefore, the application of GNNs to huge graphs presents a challenge. Although PinSAGE \cite{PinSAGE} succeeded in applying GNNs to a Web-scale network using MapReduce, their method still requires massive computational resources. There are several node sampling techniques for reducing GNN computation, such as GraphSAGE \cite{GraphSAGE}, FastGCN \cite{FastGCN}, and LADIES \cite{LADIES}, which are effective in practice. However, these techniques either are not theoretically guaranteed in terms of approximation error or incur at least a linear time computation cost.

In this study, we considered the problem of approximating the embedding of \emph{one} node using GNNs \emph{in constant time} with maximum precision.
For example, let us consider the problem of predicting whether a user of a social networking service clicks an advertisement using GNNs in real time (i.e., when the user accesses the service). The exact computation may be prohibitive because a user may have many neighbors. To make matters worse, neighboring users tend to have much more friends because users with many friends have more chances to receive a link than users with few friends. Therefore, the average degree of node GNNs with two or more layers process is higher than the average degree of the input graph.
Another example is building a browser add-on that detects malicious web pages using GNNs, where a node represents a web page and an edge represents a hyperlink. In that case, we cannot pre-compute embeddings of all the web pages because the entire graph (the WWW graph) is massive. There are two obstacles to compute the embedding on the fly. First, a page may contain many links as in the previous example. Second, retrieving contents of neighboring web pages is expensive due to communication costs. Accessing too many web pages at once may be certified as a DOS attack. Therefore, we cannot use the information of many neighbors. A natural countermeasure is sampling neighboring nodes to reduce the computational cost \cite{GraphSAGE}. However, it may lose much information and degrade much performance especially when the node has many neighbors.

We analyze the neighbor sampling technique to show that only a constant number of samples is needed to guarantee the approximation error for Message Passing Neural Networks \cite{MPNNs} including GraphSAGE \cite{GraphSAGE}, GATs \cite{GAT}, and GCNs \cite{GCN}. It should be noted that the neighbor sampling technique was introduced as a heuristic method originally, and no theoretical guarantees were provided. We prove PAC learning-like bounds of the approximation errors of neighbor sampling. Given an error tolerance $\varepsilon$ and confidence probability $1 - \delta$, our analysis shows that the following estimates can be computed in constant time, which is completely independent of the number of the nodes, edges, and neighbors.
\begin{itemize}
    \item The estimate $\hat{\boldz}_v$ of the exact embedding $\boldz_v$ of a node $v$, such that $\textnormal{Pr}[\|\hat{\boldz}_v - \boldz_v\|_2 \ge \varepsilon] \le \delta$.
    \item The estimate $\widehat{\frac{\partial \boldz_v}{\partial \boldtheta}}$ of the exact gradient $\frac{\partial \boldz_v}{\partial \boldtheta}$ of the embedding $\boldz_v$ with respect to the network parameters $\boldtheta$ such that $\textnormal{Pr}[\|\widehat{\frac{\partial \boldz_v}{\partial \boldtheta}} - \frac{\partial \boldz_v}{\partial \boldtheta}\|_F \ge \varepsilon] \le \delta$.
\end{itemize}
This result enables us to deal with graphs irrespective of their size (e.g., the WWW graph). We demonstrate that the time complexity is optimal when $L = 1$ with respect to the error tolerance $\varepsilon$. Our analysis can be applied to the prediction setting by considering the prediction problem as $1$-dimensional embedding in $[0, 1]$. In that case, the prediction with the approximation is correct if the exact computation predicts correctly with margin $\epsilon$.

In addition to presenting positive results, we show that some existing GNNs, including the original GraphSAGE, cannot be approximated in constant time by any algorithm. These results indicate that the constant-time approximation is not trivial but characteristics of certain GNN architectures.

Furthermore, in addition to providing guarantees of approximation errors, our analysis also reveals which information each GNN model gives importance to in the light of computational complexity. The GNN models that can be approximated in constant time do not use the fine-grained information of the input graph, whereas the GNN architectures that cannot be approximated in constant time do use all the information of the input graph. These observations provide theoretical characteristics of GNN architectures. \newpage

\noindent {\bf Contributions:} 
\begin{itemize}
    \item We analyze the neighbor sampling for GraphSAGE, graph attention networks (GATs), and GCNs to provide theoretical justification. Our analysis shows that the complexity is completely independent of the number of nodes, edges, and neighbors.
    \item We show that some existing GNNs, including the original GraphSAGE, cannot be approximated in constant time by any algorithm.
    \item We empirically validate our theorems using synthetic and real-world datasets.
\end{itemize}

\newcolumntype{C}{>{\centering\arraybackslash}p{0.8in}}
\begin{table*}[tb]
\small
    \caption{\cmark~ indicates that \emph{neighbor sampling} approximates the network in constant time. \xmark~ indicates that \emph{no algorithm} can approximate the network in constant time. \cmark~ in the Gradient column indicates that the error between the gradient of the approximated embedding and that of the exact embedding is also theoretically bounded. \cmark$^*$~ requires an additional condition to approximate it in constant time.}
    \centering
    \begin{tabular}{lCCccc} \toprule
    & \multicolumn{2}{c}{GATs, GraphSAGE-\{GCN, mean\}} & GraphSAGE-pool & \multicolumn{2}{c}{GCNs} \\
    \cmidrule(lr{1.0em}){2-3} \cmidrule(lr{1.0em}){5-6}
    Activation & Embedding & Gradient & & Embedding & Gradient \\ \midrule
    \multirow{2}{*}{sigmoid / tanh} & \cmark & \cmark & \xmark & \cmark$^*$ & \cmark$^*$ \\
    & Thm. \ref{inference} & Thm. \ref{gradient} & Thm. \ref{pool} & Thm. \ref{inference} & Thm. \ref{gradient} \rule[-2mm]{0mm}{2mm} \\
    \multirow{2}{*}{ReLU} & \cmark & \xmark & \xmark & \cmark$^*$ & \xmark \\
    & Thm. \ref{inference} & Thm. \ref{relu} & Thm. \ref{pool} & Thm. \ref{inference} & Thm. \ref{relu} \rule[-2mm]{0mm}{2mm} \\
    \multirow{2}{*}{ReLU + normalization} & \xmark & \xmark & \xmark & \xmark & \xmark \\
    & Thm. \ref{normalization} & Thm. \ref{normalization} & Thm. \ref{pool} & Thm. \ref{normalization} & Thm. \ref{normalization} \\ \bottomrule
    \end{tabular}
    \label{tab: summary}
\end{table*}

\section{Related Work}
\label{related}

\subsection{Graph Neural Networks (GNNs)}
Originally, GNN-like models were proposed in the chemistry field \cite{sperduti1997supervised, baskin1997neural}. Gori et al. \cite{gori} and Scarselli et al. \cite{scarselli} proposed graph learning models and named them graph neural networks (GNNs), which recursively apply the propagation function until convergence to obtain node embeddings. Bruna et al. \cite{bruna2013spectral} and Defferrard et al. \cite{ChebyNet} took advantage of graph spectral analysis and graph signal processing to construct GNN models.
GCNs \cite{GCN} approximate a spectral model by linear functions with respect to the graph Laplacian and reduced spectral models to spatial models. Gilmer et al. \cite{MPNNs} proposed message passing neural networks (MPNNs), a general framework of GNNs using the message passing mechanism. GATs \cite{GAT} improve the performance of GNNs greatly by incorporating the attention mechanism. With the advent of GATs, various GNN models with the attention mechanism have been proposed  \cite{KGCN, GENI}. 

GraphSAGE \cite{GraphSAGE} is a GNN model which employs neighbor sampling to reduce the computational costs of training and inference.
Neighbor sampling enables GraphSAGE to deal with large graphs.
However, neighbor sampling was introduced without any theoretical guarantee, and the number of samples is chosen empirically.
An alternative computationally efficient GNN would be FastGCN \cite{FastGCN}, which employs layer-wise random node sampling to speed up training and inference. Huang et al. \cite{huang2018adaptive} further improved FastGCN by using an adaptive node sampling technique to reduce the variance of estimators. By virtue of the adaptive sampling technique, it reduces the computational costs and outperforms neighbor sampling in terms of classification accuracy and convergence speed. Note that the layer-wise sampling is designed for mini-batch training and is not suitable for our setting, where we make a prediction for a single node.
Chen et al. \cite{chen2018stochastic} proposed an alternative neighbor sampling technique, which uses historical activations to reduce the estimator variance. Additionally, it can achieve zero variance after a certain number of iterations. However, because it uses the same sampling technique as GraphSAGE to obtain the initial solution, the approximation error is not theoretically bounded until the $\Omega(n)$-th iteration.  ClusterGCN \cite{ClusterGCN} first clusters nodes into dense blocks and then aggregates node features within each block. LADIES \cite{LADIES} samples neighboring nodes using importance sampling for each layer to improve efficiency while keeping variance small. However, neither LADIES \cite{LADIES} nor ClusterGCN \cite{ClusterGCN} has a theoretical guarantee on the approximation error. Overall, the existing sampling techniques are effective in practice. However, they either are not theoretically guaranteed in terms of approximation error or incur at least a linear time computation cost to calculate the embedding of a node and its gradients. In this paper, we analyze the neighbor sapling method and derive a constant time complexity. We focus on neighbor sampling owing to its simplicity. Extending our analysis to other sampling methods, including layer-wise sampling \cite{FastGCN, huang2018adaptive}, is an important future direction.

\subsection{Sublinear Time Algorithms}
Sublinear time algorithms were originally proposed for property testing \cite{rubinfeld1996robust}. Sublinear property testing algorithms determine whether the input has some property $\pi$ or the input is sufficiently far from property $\pi$ with high probability in sublinear time with respect to the input size. Sublinear time approximation algorithms are an additional type of sublinear time algorithms. More specifically, they calculate a value sufficiently close to the exact value with high probability in sublinear time. Constant time algorithms are a subclass of sublinear time algorithms. They operate not only in sublinear time with respect to the input size but also in constant time. The proposed algorithm is classified as a constant time approximation algorithm.

The examples of sublinear time approximation algorithms include minimum spanning tree in metric space \cite{czumaj2004estimating} and minimum spanning tree with integer weights \cite{chazelle2005approximating}. Parnas et al. \cite{parnas2007approximating} proposed a method to convert distributed local algorithms into constant time approximation algorithms. In their paper, they proposed a method to construct constant time algorithms for the minimum vertex cover problem and dominating set problem. Nguyen et al. \cite{nguyen2008constant} and Yoshida et al. \cite{yoshida2009improved} improved the complexities of these algorithms.
A classic example of sublinear time algorithms related to machine learning includes clustering 
\cite{indyk1999sublinear, mishra2001sublinear}.
Examples of recent studies in this stream include constant time approximation of the minimum value of quadratic functions \cite{hayashi2016minimizing} and constant time approximation of the residual error of the Tucker decomposition \cite{hayashi2017fitting}. Hayashi and Yoshida adopted simple sampling strategies to obtain a theoretical guarantee, as we did in our study. In this paper, we provide a theoretical guarantee for the approximation of GNNs within constant time for the first time.

\section{Background}
\subsection{Notations}
Let $G$ the input graph, $\mathcal{V} = \{1, 2, \dots, n\}$ the set of nodes, $n = |\mathcal{V}|$ the number of nodes, $\mathcal{E}$ the set of edges, $m = |\mathcal{E}|$ the number of edges, $\textrm{deg}(v)$ the degree of node $v$, $\mathcal{N}(v)$ the set of neighbors of a node $v$, $\boldx_v \in \mathbb{R}^{d_0}$ the feature vector associated to a node $v \in \mathcal{V}$, and $\boldX = (\boldx_1, \boldx_2 \dots, \boldx_n)^\top \in \mathbb{R}^{n \times d_0}$ the stacked feature vectors, and let $^\top$ denote the matrix transpose.

\begin{algorithm}[tb]
\caption{$\mathcal{O}_z$: Exact embedding}
\label{algo:exact}
\begin{algorithmic}[1]
\REQUIRE Graph $G = (\mathcal{V}, \mathcal{E})$; Features $\boldX \in \mathbb{R}^{n \times d_{0}}$; Node index $v \in \mathcal{V}$; Model parameters $\boldtheta$.
\ENSURE Exact embedding $\boldz_v$
\STATE $\boldz_i^{(0)} \leftarrow \boldx_i ~(\forall i \in \mathcal{V})$
\STATE \textbf{for} $l \in \{1, \dots, L\}$ \textbf{do} \hspace{0.02in} \textbf{for} $i \in \mathcal{V}$ \textbf{do}
\STATE \hspace{0.1in} $\boldh_i^{(l)} \leftarrow \sum_{u \in \mathcal{N}(i)} M_{liu}(\boldz^{(l-1)}_i, \boldz^{(l-1)}_u, \bolde_{iu}, \boldtheta)$
\STATE \hspace{0.1in} $\boldz_i^{(l)} \leftarrow U_l(\boldz_i^{(l-1)}, \boldh_i^{(l)}, \boldtheta)$
\STATE \textbf{end for} \hspace{0.02in} \textbf{end for}
\STATE \textbf{return} $\boldz^{(L)}_v$
\end{algorithmic}
\end{algorithm}

\subsection{Node Embedding Model}
We consider the node embedding problem using GNNs with the MPNN framework \cite{MPNNs}. This framework includes many GNN models, such as GraphSAGE and GCNs. Algorithm \ref{algo:exact} shows the algorithm of MPNNs. We refer to the final embedding $\boldz_v^{(L)}$ as $\boldz_v$.
The aim of this study is to show that it is possible to approximate the embedding vector $\boldz_v$ and gradients $\frac{\partial \boldz_v}{\partial \boldtheta}$ in constant time with the given model parameters $\boldtheta$ and node $v$. 

\subsection{Examples of Models}

We introduce GraphSAGE-GCN, GraphSAGE-mean, GraphSAGE-pool, the graph convolutional networks (GCNs), and the graph attention networks (GATs) for completeness of our paper. 

\vspace{.05in}
\noindent \textbf{GraphSAGE-GCN} \cite{GraphSAGE}: The message function and the update function of this model are

\begin{align*}
    M_{lvu}(\boldz_v, \boldz_u, \bolde_{vu}, \boldtheta) &= \frac{\boldz_u}{\text{deg}(v)},  \\
    U_{l}(\boldz_v, \boldh_v, \boldtheta) &= \sigma(\boldW^{(l)} \boldh_v),
\end{align*}
where $\boldW^{(l)}$ is a parameter matrix and $\sigma$ is an activation function such as sigmoid and ReLU. GraphSAGE-GCN includes the center node itself in the set of adjacent nodes (i.e., $\mathcal{N}(v) \leftarrow \mathcal{N}(v) \cup \{ v \}$).

\vspace{.05in}
\noindent \textbf{GraphSAGE-mean} \cite{GraphSAGE}: The message function and the update function of this model are

\begin{align*}
    M_{lvu}(\boldz_v, \boldz_u, \bolde_{vu}, \boldtheta) &= \frac{\boldz_v}{\text{deg}(v)}, \\
    U_{l}(\boldz_v, \boldh_v, \boldtheta) &= \sigma(\boldW^{(l)} [\boldz_v, \boldh_v]),
\end{align*}
where $[\cdot]$ denotes vertical concatenation.

\vspace{.05in}
\noindent \textbf{GraphSAGE-pool} \cite{GraphSAGE}: We do not formulate GraphSAGE-pool using the message function and the update function because it takes maximum instead of summation. The model of GraphSAGE-pool is

\[ \boldz_v^{(l)} = \max(\{\sigma(\boldW^{(l)} \boldz_u^{(l-1)} + \boldb) \mid u \in \mathcal{N}(v)\}). \]

\vspace{.05in}
\noindent \textbf{GCNs} \cite{GCN}: The message function and the update function of this model are

\begin{align*}
    M_{lvu}(\boldz_v, \boldz_u, \bolde_{vu}, \boldtheta) &= \frac{\boldz^{(l)}}{\sqrt{\text{deg}(v) \text{deg}(u)}}, \\
    U_{l}(\boldz_v, \boldh_v, \boldtheta) &= \sigma(\boldW^{(l)} \boldh_v).
\end{align*}

\vspace{.05in}
\noindent \textbf{GATs} \cite{GAT}: The message function and the update function of this model are

\begin{align*}
    \alpha_{vu}^{(l)} &= \frac{\exp (\textsc{LeakyReLU}(\bolda^{(l) \top} [\boldW^{(l)} \boldz_v, \boldW^{(l)} \boldz_u]))}{\sum_{u' \in \mathcal{N}(v)} \exp (\textsc{LeakyReLU}(\bolda^{(l) \top} [\boldW^{(l)} \boldz_v, \boldW^{(l)} \boldz_{u'}]))}, \\
    M_{lvu}(\boldz_v, \boldz_u, \bolde_{vu}, \boldtheta) &= \alpha_{vu}^{(l)} \boldz_u, \\
    U_{l}(\boldz_v, \boldh_v, \boldtheta) &= \sigma(\boldW^{(l)} \boldh_v).
\end{align*}

Technically, MPNNs do not include the above formulation of GATs because it uses embeddings of other neighboring nodes to calculate the attention value $\alpha_{vu}$. However, we can apply the same argument as MPNNs to GATs, and neighbor sampling can approximate GATs in constant time as other MPNNs. 

\subsection{Problem Formulation}
Constant time algorithms may sound tricky for unfamiliar users because reading the input takes at least linear time. The browser add-on example we introduced in the introduction makes it easier to understand how constant time algorithms work. The WWW graph is so massive that we cannot know the entire structure of the WWW graph. Nonetheless, we can run GNNs on the WWW graph without knowing the entire graph by retrieving the contents of pages in an on-demand manner.

In the design of constant time algorithms, we need to specify a means by which they can access the input because they cannot read the entire input. We follow the standard convention of sublinear time algorithms \cite{parnas2007approximating, nguyen2008constant}. We model an algorithm as an oracle machine that can generate queries regarding the input and we measure the complexity by query complexity.
Algorithms can access the input only by querying the following oracles: (1) $\mathcal{O}_{\textrm{deg}}(v)$: the degree of node $v$, (2) $\mathcal{O}_G(v, i)$: the $i$-th neighbor of node $v$, and (3) $\mathcal{O}_{\textrm{feature}}(v)$: the feature of node $v$.
We assume that an algorithm can query the oracles in constant time per query. 

Formally, given a node $v$, we compute the following functions with the least number of oracle accesses: (1) $\mathcal{O}_{\boldz}(v)$: the embedding $\boldz_v$ and (2) $\mathcal{O}_{g}(v)$: the gradients of parameters $\frac{\partial \boldz_v}{\partial \boldtheta}$.
However, the exact computation of $\mathcal{O}_{z}$ and $\mathcal{O}_{g}$ requires at least $\text{deg}(v)$ queries to aggregate the features from the neighbor nodes. Besides, the average degree of nodes GNNs process is higher than the average degree of the input graph, as we pointed out in the introduction. Thus, it is computationally expensive to execute the algorithm for a large and dense network, which motivates us to make the following approximations.
\begin{itemize}
    \item $\hat{\mathcal{O}}_{z}(v, \varepsilon, \delta)$: an estimate $\hat{\boldz}_v$ of $\boldz_v$ such that $\textnormal{Pr}[\|\hat{\boldz}_i - \boldz_i\|_2 \ge \varepsilon] \le \delta$,
    \item $\hat{\mathcal{O}}_{g}(v, \varepsilon, \delta)$: an estimate $\widehat{\frac{\partial \boldz_v}{\partial \boldtheta}}$ of $\frac{\partial \boldz_v}{\partial \boldtheta}$ such that $\textnormal{Pr}[\|\widehat{\frac{\partial \boldz_v}{\partial \boldtheta}} - \frac{\partial \boldz_v}{\partial \boldtheta}\|_{F} \ge \varepsilon] \le \delta$,
\end{itemize}
where $\varepsilon > 0$ is the error tolerance, $1 - \delta$ is the confidence probability, and $\|\cdot\|_2$ and $\|\cdot\|_F$ are the Euclidean and Frobenius norm, respectively.
Under the fixed model structure (i.e., the number of layers $L$, the message passing functions, and the update functions), we construct an algorithm that calculates $\hat{\mathcal{O}}_z$ and $\hat{\mathcal{O}}_g$ in constant time irrespective of the number of the nodes, edges, and neighbors of the input.
However, it is impossible to construct a constant time algorithm without any assumption about the inputs, as shown in Section \ref{sec:inapproximability}. Therefore, we make the following mild assumptions.

\begin{description}
\item[Assumption 1] $\exists B \in \mathbb{R}$ s.t. $\| \boldx_i \|_2 \le B$, $\| \bolde_{iu} \|_2 \le B$, and $\| \boldtheta \|_2 \le B$.
\item[Assumption 2] $\text{deg}(i) M_{liu}$ and $U_l$ are uniformly continuous in any bounded domain.
\item[Assumption 3] (Only for gradient computation) $\text{deg}(i) D M_{liu}$ and $D U_l$ are uniformly continuous in any bounded domain, where $D$ denotes the Jacobian operator.
\end{description}
Intuitively, the first assumption is needed to bound the additive error, and the second and third assumptions are needed to prohibit the amplification of errors in the message passing phase. 
Furthermore, we derive the query complexity of neighbor sampling when the message functions and update functions satisfy the following assumptions.

\begin{description}
\item[Assumption 4] $\exists K \in \mathbb{R}$ s.t. $\text{deg}(i) M_{liu}$ and $U_l$ are $K$-Lipschitz continuous in any bounded domain.
\item[Assumption 5] (Only for gradient computation) $\exists K' \in \mathbb{R}$ s.t. $\text{deg}(i) D M_{liu}$ and $D U_l$ are $K'$-Lipschitz continuous in any bounded domain.
\end{description}

\section{Main Results}
\label{proposed}

\begin{algorithm*}[t]
\caption{$\hat{\mathcal{O}}_{z}^{(l)}$: Estimate the embedding $\boldz^{(l)}_v$}
\label{algo: constant}
\begin{algorithmic}[1]
\REQUIRE Graph $G = (\mathcal{V}, \mathcal{E})$ (as oracle); Features $\boldX \in \mathbb{R}^{n \times d_{0}}$ (as oracle); Node index $v \in \mathcal{V}$; Model parameters $\boldtheta$; Error tolerance $\varepsilon$; Confidence probability $1 - \delta$.
\ENSURE Approximation of the embedding $\boldz^{(l)}_v$
\STATE $\mathcal{S}^{(l)} \leftarrow$ sample $r^{(l)}(\varepsilon, \delta)$ neighbors of $v$ with uniform random with replacement.
\STATE $\hat{\boldh}_v^{(l)} \leftarrow \begin{cases}
\frac{\mathcal{O}_{\textrm{deg}}(v)}{r^{(l)}} \sum_{u \in \mathcal{S}^{(l)}} M_{lvu}(\mathcal{O}_{\text{feature}}(v), \mathcal{O}_{\text{feature}}(u), \bolde_{iu}, \boldtheta) \hfill \quad (l = 1)\\
\frac{\mathcal{O}_{\textrm{deg}}(v)}{r^{(l)}} \sum_{u \in \mathcal{S}^{(l)}} M_{lvu}(\hat{\mathcal{O}}_{z}^{(l-1)}(v \leftarrow v), \hat{\mathcal{O}}_{z}^{(l-1)}(v \leftarrow u), \bolde_{iu}, \boldtheta) \hfill \quad (l > 1)
\end{cases}$
\STATE $\hat{\boldz}_v^{(l)} \leftarrow U_l(\hat{\mathcal{O}}_{z}^{(l-1)}(v \leftarrow v), \hat{\boldh}_v^{(l)}, \boldtheta)$ if $l > 1$ otherwise $U_l(\mathcal{O}_{\text{feature}}(v), \hat{\boldh}_v^{(l)}, \boldtheta)$
\STATE \textbf{return} $\hat{\boldz}_i$
\end{algorithmic}
\end{algorithm*}

\subsection{Constant Time Embedding Approximation}
We describe the construction of a constant time approximation algorithm based on neighbor sampling, which approximates the embedding $\boldz_v$ with an absolute error of at most $\varepsilon$ and probability $1 - \delta$.
We recursively constructed the algorithm layer by layer by sampling $r^{(l)}$ neighboring nodes in layer $l$. We refer to the algorithm that calculates the estimate of the embeddings in the $l$-th layer $\boldz^{(l)}$ as $\hat{\mathcal{O}}_{z}^{(l)} (l = 1, \dots, L)$. Algorithm \ref{algo: constant} presents the pseudo code. Here, $\hat{\mathcal{O}}_{z}^{(l-1)}(v \leftarrow u)$ represents calling the function $\hat{\mathcal{O}}_{z}^{(l-1)}$ with the same parameters as the current function, except for $v$, which is replaced by $u$. In the following, we demonstrate the theoretical properties of Algorithm \ref{algo: constant}. The following theorem shows that the approximation error of Algorithm \ref{algo: constant} is bounded by $\varepsilon$ with probability $1 - \delta$. It is proved by applying Hoeffding's inequality \cite{Hoeffding} recursively. Because the number of sampled nodes depends only on $\varepsilon$ and $\delta$ and is independent of the number of the nodes, edges, and neighbors of the input, Algorithm \ref{algo: constant} operates in constant time. We stress that the only source of randomness in the analysis of Theorem \ref{inference} is the sampling distribution of neighboring nodes. This is irrelevant to the generation process of the input graph. Theorem \ref{inference} holds even if the feature distribution of the input graph is skewed and correlated to each other.

\begin{theorem} \label{inference}
For all $\varepsilon > 0, 1 > \delta > 0$, there exists $r^{(l)}(\varepsilon, \delta)~  (l = 1, \dots, L)$ such that for all inputs satisfying Assumptions 1 and 2, the following property holds true:
\[ \textnormal{Pr}[\|\hat{\mathcal{O}}_{z}(v, \varepsilon, \delta) - \boldz_v\|_2 \ge \varepsilon] \le \delta. \]
\end{theorem}

\begin{sketch}
We prove the theorem by performing mathematical induction on the number of layers $L$. First, we prove that the norms of the embeddings are bounded under the assumptions. In the base case, we can bound the sampling error by Hoeffding's inequality owing to the bounded norms. In the inductive step, the error comes from the previous layer and the sampling error in the current layer. We bound the former by the induction hypothesis and the latter by Hoeffding's inequality.
\end{sketch}

All proofs are available in Section \ref{sec: proof}. Note that the i.i.d. assumption of uniform sampling is crucial in Hoeffding's inequality used in our analysis. If a correlated sampling method holds a Hoeffding-style concentration inequality, our analyses can be extended to such a sampling method. We leave extending our analysis to other sampling methods for future work.

Next, we provide the complexity when the functions are Lipschitz continuous. This theorem shows a rough estimate of a sufficient number of the sampling size $r$. We confirm that this theoretical sampling rate is valid in the experiments.

\begin{theorem} \label{time}
Under Assumptions 1 and 4, $r^{(L)} = O(\frac{1}{\varepsilon^2} \log \frac{1}{\delta})$ and $r^{(1)}, \dots, r^{(L-1)} = O(\frac{1}{\varepsilon^2} (\log \frac{1}{\varepsilon} + \log \frac{1}{\delta}))$ are sufficient, and the query complexity of Algorithms \ref{algo: constant} is $O(\frac{1}{\varepsilon^{2L}} (\log \frac{1}{\varepsilon} + \log \frac{1}{\delta})^{L-1} \log \frac{1}{\delta})$.
\end{theorem}

\begin{sketch}
We prove the theorem by performing mathematical induction on the number of layers $L$ as in the previous theorem. Thanks to the Lipschitz assumption, we can quantitatively bound the error expansion in this case. The complete proof is available in Section \ref{sec: proof}.
\end{sketch}

Although the complexity is exponential with respect to the number of layers, this is nonetheless beneficial because the number of layers is usually small in practice. For example, the original GCN employs two layers \cite{GCN}. It is noteworthy that, although most constant time algorithms proposed in the literature also depend on some parameters exponentially, they have nonetheless been proved to be effective. For example, the constant time algorithms of Yoshida et al. \cite{yoshida2009improved} for the maximum matching problem and minimum set cover problem use $d^{O(\frac{1}{\varepsilon^2})} (\frac{1}{\varepsilon})^{O(\frac{1}{\varepsilon})}$ and $\frac{1}{\varepsilon^2} (st)^{O(s)}$ queries, respectively, where $d$ is the maximum degree and $s$ is the number of elements in a set, and each element is in at most $t$ sets. The important point here is that the complexity is completely independent of the size of the inputs, which is desirable, especially when the input size can be very large. In addition, we show that the query complexity of Algorithm \ref{algo: constant} is optimal with respect to $\varepsilon$ if the number of layers is one. In other words, a one-layer model cannot be approximated in $o(\frac{1}{\varepsilon^2})$ time by any algorithm.  

\begin{theorem} \label{opt}
Under Assumptions 1 and 4 and $L = 1$, the time complexity of Algorithm \ref{algo: constant} in Theorem \ref{time} is optimal with respect to the error tolerance $\varepsilon$.
\end{theorem}

\begin{sketch}
We reduce the problem to estimating a parameter of a simple distribution. We prove it impossible to determine the parameter in $o(\varepsilon^2)$ queries by Chazelle's Lemma \cite{chazelle2005approximating}. The complete proof is available in Section \ref{sec: proof}.
\end{sketch}

The optimality when $L \ge 2$ is an open problem.

\subsection{Constant Time Gradient Approximation.}
We show that the neighbor sampling can guarantee the approximation errors of the gradient of embeddings with respect to the model parameters. Let $\frac{\partial \boldz_v}{\partial \boldtheta}$ be the gradient of the embedding $\boldz_v$ with respect to the model parameter $\boldtheta$, i.e., $(\frac{\partial \boldz_v}{\partial \boldtheta})_{ijk} = \frac{\partial \boldz_{vi}}{\partial \boldtheta_{jk}}$. The following theorem shows that an estimate of the gradient of the embedding with respect to parameters with an absolute error of at most $\varepsilon$ and probability $1 - \delta$ can be calculated by running $\hat{\mathcal{O}}^{(L)}_{z}(v, \varepsilon, \delta)$ and calculating the gradient of the obtained estimate of the embedding.

\begin{theorem} \label{gradient}
For all $\varepsilon > 0, 1 > \delta > 0$, there exists $r^{(l)}(\varepsilon, \delta)~  (l = 1, \dots, L)$ such that for all inputs satisfying Assumptions 1, 2, and 3, the following property holds true:
\[ \textnormal{Pr}[\|\widehat{\frac{\partial \boldz^{(L)}_v}{\partial \boldtheta}} - \frac{\partial \boldz^{(L)}_v}{\partial \boldtheta}\|_F \ge \varepsilon] \le \delta, \]
where $\widehat{\frac{\partial \boldz^{(L)}_v}{\partial \boldtheta}}$ is the gradient of $\hat{\boldz}^{(L)}_v$, which is obtained by $\hat{\mathcal{O}}^{(L)}_{z}(v, \varepsilon, \delta)$, with respect to $\boldtheta$.
\end{theorem}

\begin{sketch}
The basic strategy is common with Theorem \ref{inference}. However, the derivation becomes more challenging and complicated due to additional terms in the backward path. The complete proof is available in Section \ref{sec: proof}.
\end{sketch}

Next, we provide the complexity when the functions are Lipschitz continuous.

\begin{theorem} \label{gtime}
Under Assumptions 1, 4, and 5, $r^{(L)} = O(\frac{1}{\varepsilon^2} \log \frac{1}{\delta})$ and $r^{(1)}, \dots, r^{(L-1)} = O(\frac{1}{\varepsilon^2} (\log \frac{1}{\varepsilon} + \log \frac{1}{\delta}))$ are sufficient, and the gradient of the embedding with respect to parameters can be approximated with an absolute error of at most $\varepsilon$ and probability $1 - \delta$ in $O(\frac{1}{\varepsilon^{2L}} (\log \frac{1}{\varepsilon} + \log \frac{1}{\delta})^{L-1} \log \frac{1}{\delta})$ time.
\end{theorem}

\begin{sketch}
The Lipschitz assumption enables to quantitatively bound the error expansion. The proof is available in Section \ref{sec: proof}.
\end{sketch}

\subsection{Constant Time Approximation of Graph Attention Networks}

Technically speaking, MPNNs do not include GATs, because these networks use embeddings of other neighboring nodes to calculate the attention value.  However, our analysis can be naturally extended to GATs, and we can approximate GATs in constant time by neighbor sampling. This is surprising because GATs may pay considerable attention to some nodes, and uniform sampling may fail to sample these nodes. However, it can be shown that our assumptions, which do not seem related to this issue, prohibit this situation. 

To be precise, the following proposition holds true.
\begin{proposition} \label{prop: GAT}
If Assumption 1 holds true and $\sigma$ is Lipschitz continuous, and we take $r^{(L)} = O(\frac{1}{\varepsilon^2} \log \frac{1}{\delta})$ and $r^{(1)}, \dots, r^{(L-1)} = O(\frac{1}{\varepsilon^2} (\log \frac{1}{\varepsilon} + \log \frac{1}{\delta}))$ samples, and let
\begin{align*}
\hat{\boldz}^{(0)}_v &= \boldz^{(0)}_v, \\
\hat{\alpha}_{vu}^{(l)} &= \frac{\exp (\textsc{LeakyReLU}(\bolda^{(l) \top} [\boldW^{(l)} \hat{\boldz}^{(l-1)}_v, \boldW^{(l)} \hat{\boldz}^{(l-1)}_u]))}{\sum_{u' \in \mathcal{S}^{(l)}} \exp (\textsc{LeakyReLU}(\bolda^{(l) \top} [\boldW^{(l)} \hat{\boldz}^{(l-1)}_v, \boldW^{(l)} \hat{\boldz}^{(l-1)}_{u'}]))}, \\
\hat{\boldh}_v^{(l)} &= \sum_{u \in \mathcal{S}^{(l)}} \hat{\alpha}_{vu}^{(l)} \hat{\boldz}^{(l-1)}_u, \\
\hat{\boldz}_v^{(l)} &= U_{l}(\hat{\boldz}^{(l-1)}_v, \hat{\boldh}^{(l)}_v, \boldtheta).
\end{align*}
Then, the following property holds true.
\[ \textnormal{Pr}[ \|\boldz^{(L)}_v - \hat{\boldz}^{(L)}_v \|_2 \ge \varepsilon] < \delta. \]
\end{proposition}

\begin{sketch}
The main difference between GCN and GAT is that GAT can put adaptive weights to neighboring nodes. Although large weights can blow up the approximation error, the magnitude of the weights in GAT is bounded under the assumptions. Therefore, the embeddings are also bounded, and we can adopt the same strategy as Theorem \ref{inference}. The proof is available in Section \ref{sec: proof}.
\end{sketch}

\section{Inapproximability}
\label{sec:inapproximability}
In this section, we show that some existing GNNs cannot be approximated in constant time. The theorems state that these models cannot be approximated in constant time by either neighbor sampling or \emph{any other algorithm}. In other words, for any algorithm that operates in constant time, there exists an error tolerance $\varepsilon$, a confidence probability $1 - \delta$, and a counter example input such that the approximation error for the input is more than $\varepsilon$ with probability $\delta$. This indicates that the application of an approximation method to these models requires close supervision because the obtained embedding may be significantly different from the exact embedding. These results also indicate that the positive results we have shown so far are not void but non-trivial properties.

The following proposition indicates that Assumption 1 is necessary for constant-time approximation.

\begin{proposition} \label{bound}
If $\|\boldx_i\|_2$ or $\|\boldtheta\|_{F}$ is not bounded, even under Assumption 2, the embeddings of GraphSAGE-GCN cannot be approximated with arbitrary precision and probability in constant time.
\end{proposition}

\begin{sketch}
We prove this proposition by constructing a concrete counter-example. The proof is available in Section \ref{sec: proof}.
\end{sketch}

All proofs in this section are based on constructing counter examples and available in Section \ref{sec: proof}.
The following proposition shows that the original GraphSAGE-GCN cannot be approximated in constant time.

\begin{proposition} \label{normalization}
Even under Assumption 1, the embeddings and gradients of GraphSAGE-GCN with ReLU activation and normalization cannot be approximated with arbitrary precision and probability in constant time.
\end{proposition}

The following proposition indicates that the activation function is important for constant-time approximability because the gradient of GraphSAGE-GCN with sigmoid activation \emph{can} be approximated within constant time by neighbor sampling (Theorem \ref{gradient}).

\begin{proposition} \label{relu}
Even under Assumptions 1 and 2, the gradients of GraphSAGE-GCN with ReLU activation cannot be approximated with arbitrary precision and probability in constant time.
\end{proposition}

The following two theorems state that GraphSAGE-pool and GCN cannot be approximated in constant time even under Assumptions 1, 2, and 3.

\begin{proposition} \label{pool}
Even under Assumptions 1, 2, and 3, the embeddings of GraphSAGE-pool cannot be approximated with arbitrary precision and probability in constant time.
\end{proposition}

\begin{proposition} \label{gcn}
Even under Assumptions 1, 2, and 3, the embeddings of GCN cannot be approximated with arbitrary precision and probability in constant time.
\end{proposition}

\subsection{Constant Time Approximation for GCNs}
Because of the inapproximability theorems, GCNs cannot be approximated in constant time. However, GCNs can be approximated in constant time if the input graph satisfies the following property.

\begin{description}
\item[Assumption 6] There exists a constant $C \in \mathbb{R}$ such that, for any input graph $G = (\mathcal{V}, \mathcal{E})$ and node $v, u \in \mathcal{V}$, the ratio of $\text{deg}(v)$ to $\text{deg}(u)$ is at most $C$ (i.e., $\text{deg}(v) / \text{deg}(u) \le C$).
\end{description}

Assumption 6 prohibits input graphs that have a skewed degree distribution. GCNs require Assumption 6 because the norm of the embedding is not bounded, and the influence of anomaly nodes with low degrees is significant without it.

\begin{proposition} \label{prop: GCN}
If Assumptions 1 and 6 hold true, $\sigma$ is Lipschitz continuous, and we take $r^{(L)} = O(\frac{1}{\varepsilon^2} \log \frac{1}{\delta})$ and $r^{(1)}, \dots, r^{(L-1)} = O(\frac{1}{\varepsilon^2} (\log \frac{1}{\varepsilon} + \log \frac{1}{\delta}))$ samples, and let
\begin{align*}
\hat{\boldz}^{(0)}_v &= \boldz^{(0)}_v, \\
\hat{\boldh}_v^{(l)} &= \frac{\text{deg}(v)}{|\mathcal{S}^{(l)}|} \sum_{u \in \mathcal{S}^{(l)}} \frac{\hat{\boldz}^{(l-1)}_u}{\sqrt{\text{deg}(v) \text{deg}(u)}}, \\
\hat{\boldz}_v^{(l)} &= U_{l}(\hat{\boldz}^{(l-1)}_v, \hat{\boldh}^{(l)}_v, \boldtheta).
\end{align*}
Then, the following property holds true.
\[ \textnormal{Pr}[ \|\boldz^{(L)}_v - \hat{\boldz}^{(L)}_v \|_2 \ge \varepsilon] < \delta. \]
\end{proposition}

\begin{sketch}
Although the norms of node embeddings of GCN may increase arbitrarily, they are bounded under Assumption 6. The same strategy as Theorem 1 can be used with a slight modification. The proof is available in Section \ref{sec: proof}.
\end{sketch}

It should be noted that the GraphSAGE-pool cannot be approximated in constant time even under Assumption 6.

\subsection{Discussion on Negative Results}
These negative results reveal which information each GNN model gives importance to. For example, suppose there is a high-degree node that neighbors anomaly low-degree nodes in the input graph. GraphSAGE-GCN can be approximated in constant time even in this case according to Theorem \ref{inference}, but GCNs cannot be approximated according to Proposition \ref{gcn}. GraphSAGE-GCN can be approximated in constant time because the exact computation of this model can be estimated by only a fraction of the input graph. This fraction hardly contains the anomaly nodes. Nevertheless, the estimation from the fraction is accurate. It means that the anomaly nodes do not perturb the exact computation of this model. In contrast, GCNs cannot be approximated in constant time because the anomaly nodes change the exact computation drastically. Without using anomaly nodes as well as majority nodes, the estimation becomes inaccurate. These observations tell us the characteristics of these models. If the anomaly nodes are important for estimating the label (e.g., fraud detection), we should use GCNs because GCNs can take anomaly nodes into consideration in model inference, whereas GraphSAGE-GCN ignores anomalies. If we want the model to be robust to anomaly nodes, we should use GraphSAGE-GCN. These observations are valid even when we do not use approximations, though the starting point was the time complexity of approximation methods.
This type of observation can be applied to other models such as GraphSAGE-pool and GATs as well.

Besides, these results reveal the graph problems that GNNs cannot solve via the lens of time complexity. For example, let a node be positive if there exists at least one neighboring node with feature $1$, and negative otherwise. This problem is not solvable in sublinear time because a star graph with no ``$1$'' nodes and a star graph with only one ``$1$'' leaf node are counterexamples. Therefore, GraphSAGE-GCNs, GCNs, and GATs cannot solve this problem. If these models can solve this problem, we can solve this problem in constant time by neighbor sampling, which leads to a contradiction. This result means that these models cannot simulate the pooling operation. By contrast, GraphSAGE-pool can solve this problem owing to the pooling operator, and GraphSAGE-pool indeed requires at least linear time computation according to Proposition \ref{pool}.

\section{Graph Embedding}
Our analysis can be extended to graph embedding, where we embed an entire graph instead of a node. Graph embeddings can be calculated by aggregating the embeddings of all nodes \cite{MPNNs}:
\[ \boldz_G = \textsc{READOUT}(\{\boldz_i \mid i \in V\}). \]
We adopt the mean of the feature vectors of the nodes as the readout function (\textit{i.e.,} $\boldz_G = \frac{1}{n} \sum_{i \in V} \boldz_i$). We cannot calculate the embeddings of all nodes in constant time even if each calculation is done in constant time because there are $n$ nodes. We adopt the sampling strategy here as well. We sample some nodes in a uniformly random manner, compute their feature vectors in constant time using Algorithm \ref{algo: constant}, and calculate their empirical mean. The errors of sampling and Algorithm \ref{algo: constant} are bounded by Lemma \ref{hoeffding_multi} and Theorem \ref{inference}, respectively. Therefore, we sample a sufficiently large (but independent of the graph size) number of nodes and call Algorithm \ref{algo: constant} with sufficiently small $\varepsilon$ and $\delta$. Then, the estimate is arbitrarily close to the exact embedding of $G$ with an arbitrary probability.

\section{Experiments}
\label{experiments}

\begin{figure*}[tb]
\begin{minipage}{0.31\hsize}
\begin{center}
\includegraphics[width=\hsize]{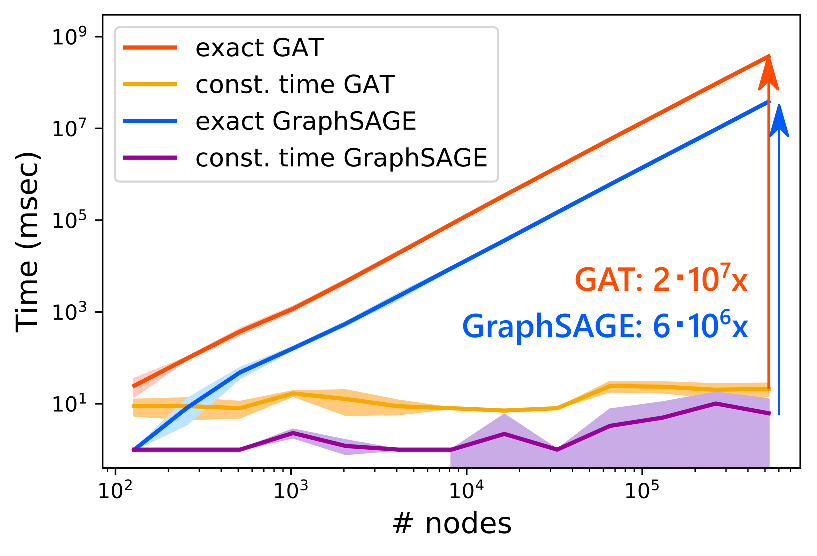}
(a) Speed (Q1).
\end{center}
\end{minipage}
\hspace{-0.05in}
\begin{minipage}{0.34\hsize}
\begin{center}
\includegraphics[width=\hsize]{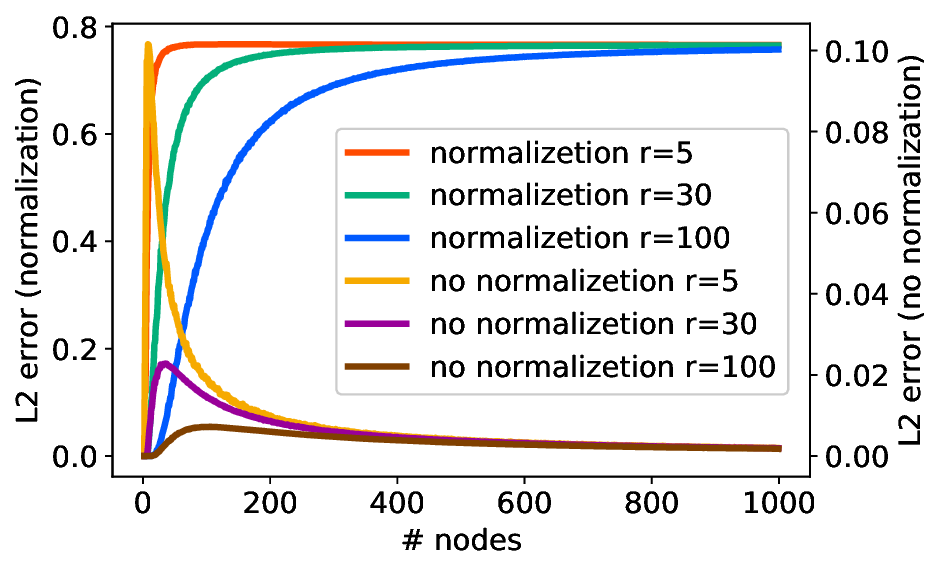}
(b) Inference (Q2).
\end{center}
\end{minipage}
\hspace{-0.05in}
\begin{minipage}{0.34\hsize}
\begin{center}
\includegraphics[width=\hsize]{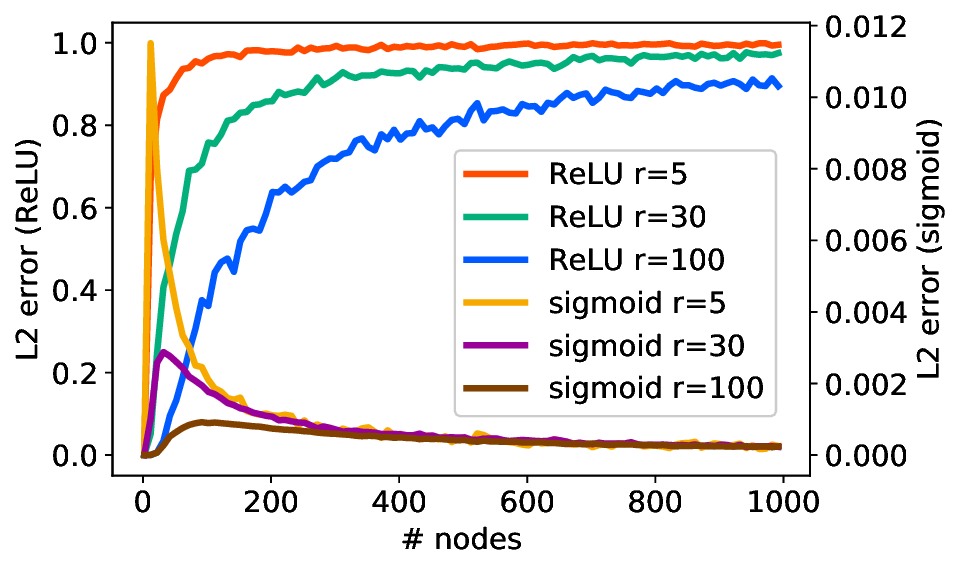}
(c) Gradient (Q3).
\end{center}
\end{minipage}
\begin{minipage}{0.305\hsize}
\begin{center}
\includegraphics[width=\hsize]{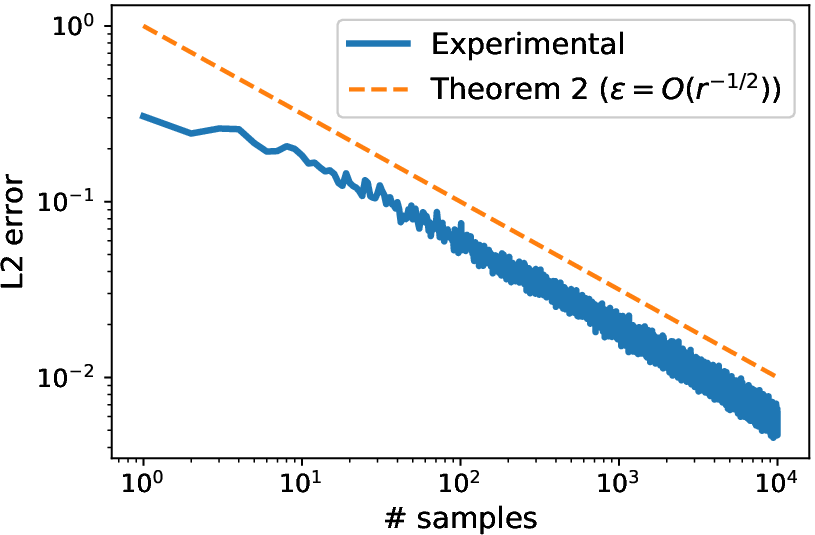}
\vspace{0.001in}
\hspace{0.15in} (d) Theoretical rate (Q4).
\end{center}
\end{minipage}
\begin{minipage}{0.305\hsize}
\begin{center}
\includegraphics[width=\hsize]{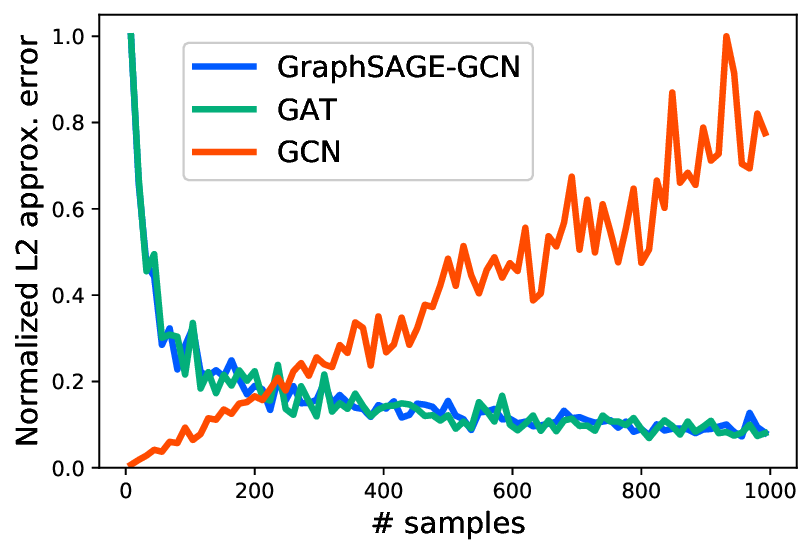}
(e) BA model (Q5).
\end{center}
\end{minipage}
\hspace{0.15in}
\begin{minipage}{0.305\hsize}
\begin{center}
\includegraphics[width=\hsize]{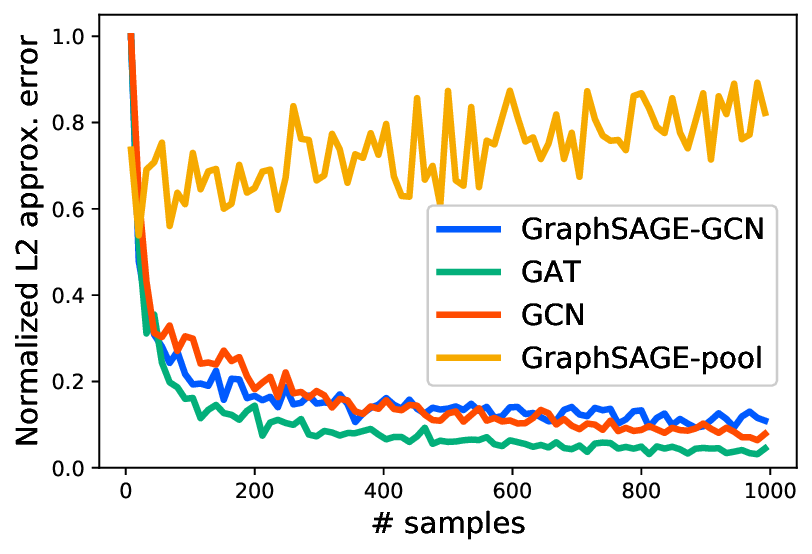}
(f) ER model (Q5).
\end{center}
\end{minipage}

\caption{Inference speed and approximation errors. Details are described in the text.}
\label{fig: plot}
\end{figure*}

We validate our theoretical results through numerical experiments. Namely, we answer the following questions through experiments.
\begin{itemize}
    \item \textbf{Q1:} How fast is the constant time approximation algorithm?
    \item \textbf{Q2:} Does the neighbor sampling accurately approximate the embeddings of GraphSAGE-GCN without normalization (Theorem \ref{inference}), whereas it cannot approximate the original one (Proposition \ref{normalization})?
    \item \textbf{Q3:} Does the neighbor sampling accurately approximate the gradients of GraphSAGE-GCN with sigmoid activation (Theorem \ref{gradient}), whereas it cannot approximate that with ReLU activation (Proposition \ref{relu})?
    \item \textbf{Q4:} Is the theoretical rate of the approximation error of Algorithm \ref{algo: constant} tight?
    \item \textbf{Q5:} Does the neighbor sampling fail to approximate GCNs when the degree distribution is skewed (Proposition \ref{gcn})? Does it succeed when the node distribution is flat (Assumption 6)?
    \item \textbf{Q6:} Does the neighbor sampling work efficiently for real data?
    \item \textbf{Q7:} How does the neighbor sampling affect downstream tasks?
\end{itemize}   

We use cliques for Q1 to Q4 for the following reasons. First, sampling is effective when the input graph is dense. Cliques are the densest graphs. We use them to highlight the effectiveness of the node sampling. Second, cliques are important structures in practice. For example, deep sets \cite{deepsets} are a popular machine learning model for sets. Equivariant deep sets can be seen as GNNs that run on cliques. Self attention layers \cite{vaswani2017attention} can also be seen as graph attention networks that run on cliques.

\subsection{Speedup Factors (Q1)}
We measure the speed of exact computation and the neighbor sampling of two-layer GraphSAGE-GCN and two-layer GATs. We initialize parameters using the i.i.d. standard multivariate normal distribution. The input graph is a clique $K_n$. We use ten-dimensional vectors from the i.i.d. standard multivariate normal distribution as the node features. We take $r^{(1)} = r^{(2)} = 100$ samples. For each method and $n = 2^7, 2^8, \dots, 2^{19}$, we run inference $10$ times and measure the average time consumption and standard deviation. The speed is evaluated with a single core of Intel Xeon CPU E5-2690. Figure \ref{fig: plot} (a) plots the speed of these methods as the number of nodes increases. This shows that the neighbor sampling is several orders of magnitude faster than the exact computation when the graph size is large.

\subsection{Effect of Normalization (Q2)}
We use the original one-layer GraphSAGE-GCN (with ReLU activation and normalization) and one-layer GraphSAGE-GCN with ReLU activation. The input graph is a clique $K_n$, the features of which are $\boldx_1 = (1, 0)^\top$ and $\boldx_i = (0, 1/n)^\top~(i \neq 1)$, and the weight matrix is an identity matrix $\boldI_2$. We use $r^{(1)} = 5, 30,$ and $100$ as the sample size. If a model can be approximated in constant time, the approximation error goes to zero as the sample size increases, even if the graph size reaches infinity. 
Figure \ref{fig: plot} (b) illustrates the approximation errors of both models. The approximation error of the original GraphSAGE-GCN converges to approximately $0.75$ even if the sample size increases. In contrast, the approximation error without normalization becomes increasingly bounded as the sample size increases. This is consistent with Theorems \ref{inference} and \ref{normalization}.

\subsection{Effect of Activation functions (Q3)}
We examine the approximation errors of the gradients using the one-layer GraphSAGE-GCN with ReLU and sigmoid activation. The input graph is a clique $K_n$, the features of which are $\boldx_1 = (1, 2)^\top$ and $\boldx_i = (1,1)^\top~(i \neq 1)$, and the weight matrix is $((-1, 1))$. We use $r^{(1)} = 5, 30,$ and $100$ as the sample size.
Figure \ref{fig: plot} (c) illustrates the approximation error of both models. The approximation error with ReLU activation converges to approximately $1.0$, even if the sample size increases. In contrast, the approximation error with sigmoid activation becomes increasingly bounded as the sample size increases. This is consistent with Theorems \ref{gradient} and \ref{relu}.

\subsection{Theoretical Rate (Q4)}
We use one-layer GraphSAGE-GCN with sigmoid activation. We initialize the weight matrix $\boldW^{(1)}$ with normal distribution and then normalize it so that the operator norm $\| \boldW^{(1)} \|_{\text{op}} $ of the matrix is equal to $1$. This satisfies Assumption 1, i.e., $\| \boldtheta \|_2 \le \sqrt{2}$. The input graph is a clique $K_n$ with $n = 40000$ nodes. We set the dimensions of intermediate embeddings as $2$, and each feature value is set to $1$ with probability $0.5$ and $-1$ otherwise. This satisfies Assumption 1, i.e., $\|\boldx_i\|_2 \le \sqrt{2}$. We compute the approximation errors of Algorithm $\ref{algo: constant}$ with different numbers of samples. Specifically, for each $r = 1, \dots, 10000$, we (1) initialize the weight matrix, (2) choose $400$ nodes, (3) calculate the exact embedding of each chosen node, (4) calculate the estimate for each chosen node with $r$ samples, i.e., $r^{(1)} = r$, and (5) calculate the approximation error of each chosen node.
Figure \ref{fig: plot} (d) illustrates the $99$-th percentile point of empirical approximation errors and the theoretical bound by Theorem \ref{time}, i.e., $\varepsilon = O(r^{-1/2})$. It shows that the approximation error decreases together with the theoretical rate. This indicates that the theoretical rate is tight. Based on these experimental and theoretical results, we can estimate a sufficient number of samples given the required precision.

\subsection{Other Architectures (Q5)}
We analyze the instances when neighbor sampling succeeds and fails for a variety of models. First, we use the Barabasi--Albert (BA) model \cite{BAmodel}. The degree distribution of the BA model follows a power law, which indicates neighbor sampling will fail to approximate GCNs (Propositions \ref{gcn} and \ref{prop: GCN}). We use ten-dimensional vectors from the i.i.d. standard multivariate normal distribution as the node features. We use two-layer GraphSAGE-GCN, GATs, and GCNs with ReLU activation. We use the same number $r$ of samples in the first and second layer, i.e., $r = r^{(1)} = r^{(2)} \in [8, 1000]$ and use graphs with $n = r^2$ nodes. Specifically, we (1) iterate $r$ from $8$ to $1000$, (2) set $n = r^2$, (3) generate 10 graphs with $n$ nodes using the BA model, (4) choose the node that has the maximum degree for each generated graph, (5) calculate the exact embeddings and its estimate for each chosen node with $r$ samples, i.e., $r^{(1)} = r^{(2)} = r$, and (6) calculate the approximation error. We use the maximum degree node in step (4) because this is the hardest case and thus shows a clear distinction between appropriate and inappropriate situations. Low degree nodes with a few neighboring nodes are easy to approximate even with an inappropriate configuration. We focus on the hardest case to illustrate the distinction clearly in the following analysis.

Figure \ref{fig: plot} (e) shows that the error of GCNs linearly increases, even if the number of samples increases while the errors of GraphSAGE-GCN and GATs gradually decrease. This is consistent with Proposition \ref{gcn}. This result indicates that the approximation of GCNs requires close supervision when the input graph is a social network because the degree distribution of a social network presents the power law as the BA model.

Next, we use the Erd\H{o}s--R\'{e}nyi (ER) model \cite{ERmodel}. It generates graphs with a flat degree distribution. We use the two-layer GraphSAGE-GCN, GATs, GCNs, and GraphSAGE-pool. The experimental process is similar to that for the BA model, but (1) we use the ER model instead of the BA model and (2) set $n = \text{floor}(r^{1.5})$ instead of $n = r^2$ to reduce the computational cost.
Figure \ref{fig: plot} (f) shows the approximation error. It shows that the errors of GraphSAGE-GCN, GATs, and GCNs gradually decrease as the number of samples increases. This is consistent with Theorem \ref{inference} and Proposition \ref{prop: GCN}. In contrast, the approximation error of GraphSAGE-pool does not decrease, even if the input graphs are generated by the ER model. This is consistent with Proposition \ref{pool}.

\subsection{Real World Datasets (Q6)}

\begin{figure}[tb] 
\centering
\includegraphics[width=\hsize]{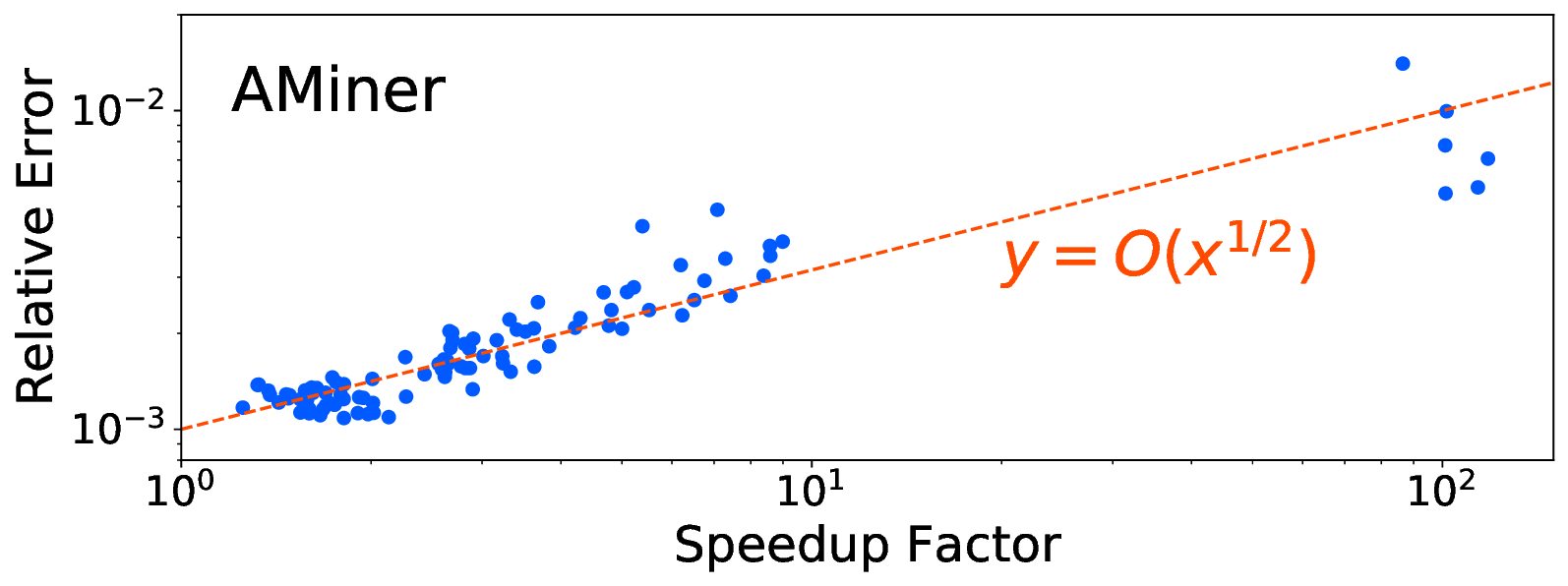}
\includegraphics[width=\hsize]{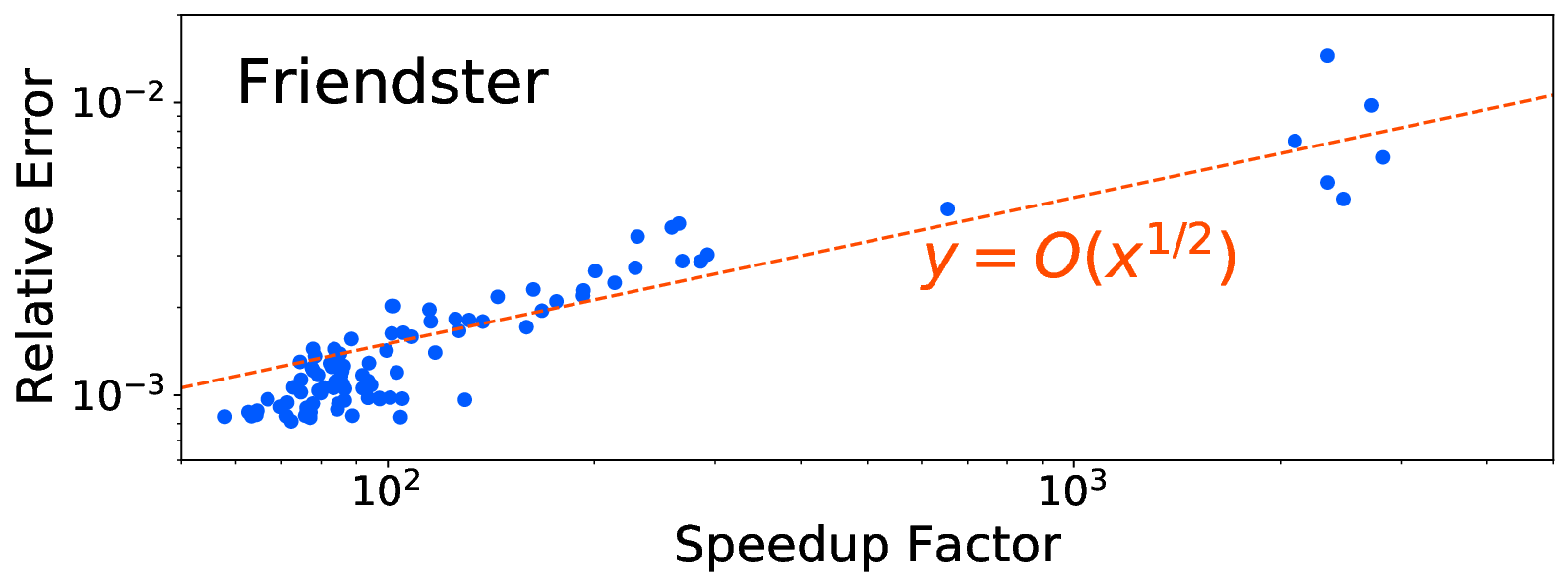}
\includegraphics[width=\hsize]{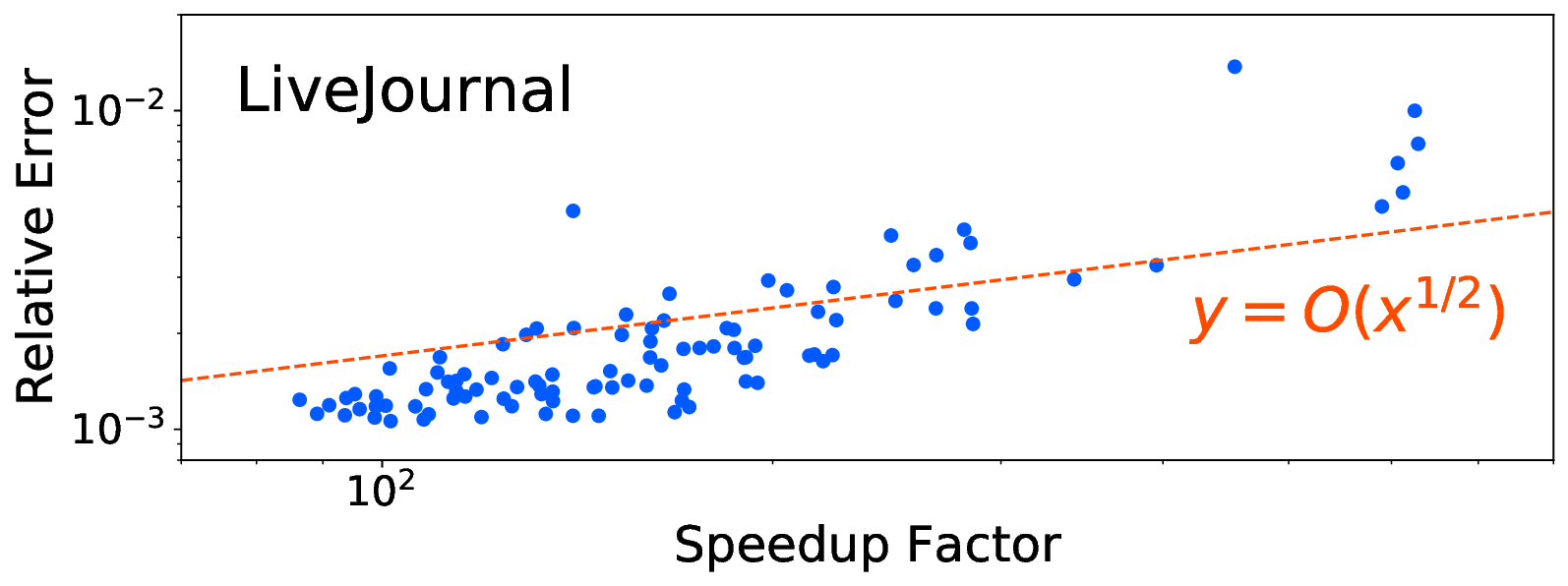}
\caption{Trade-off of speed up and relative error in real world datasets. Neighbor sampling computes embeddings a few orders of magnitude faster than exact computation within 1\% relative error.}
\label{fig: real}
\end{figure}

We assess the speed and accuracy of neighbor sampling approximation using three large-scale real world datasets: AMiner citation network \footnote{\url{https://www.aminer.cn/aminernetwork}}, Friendster social network \footnote{\url{https://snap.stanford.edu/data/com-Friendster.html}}, and LiveJournal social network \footnote{\url{https://snap.stanford.edu/data/com-LiveJournal.html}}. They contain $2092356$, $65608366$, and $3997962$ nodes, respectively. We use two-layer GraphSAGE-GCN with sigmoid activation and randomly initialize parameters by the Xavier initializer \cite{xavier}. The initial embeddings of nodes are generated from i.i.d. $128$-dimensional standard Gaussian distribution. The number of dimensions in intermediate embeddings is also $128$. For each $r = 1, \dots, 100$, we compute the exact and approximated embeddings of node with top-$10$ degrees and measure the relative error $y = \|\boldz_\text{exact} -  \boldz_\text{approx}\| / \|\boldz_\text{exact}\|$, where $\boldz_\text{exact}$ is the exact embedding and $\boldz_\text{approx}$ is the approximated embedding, and the speedup factor $x = t_\text{exact} / t_\text{approx}$, where $t_\text{exact}$ is the time consumption of the exact computation and $t_\text{approx}$ is the time consumption of the sampling method. Figure \ref{fig: real} plots these values. This shows that the approximation errors drop quickly, in particular, with rate around $O(x^{1/2})$. These results are consistent with our theoretical analyses. Based on these experimental and theoretical results, we can estimate a sufficient number of samples given a required precision.

\subsection{Affect to Downstream Tasks (Q7)}

\begin{table*}[tb]
\small
    \centering
    \caption{Node Classification performance (F1-score) with various $r$. The highest score and scores within $\pm$ std dev of the highest score are marked in \textbf{bold}. The relative errors reported in parentheses are the approximation errors of node embeddings.} 
    \begin{tabular}{lcccc} \toprule
    & Cora & Cora Full & PubMed & Citeseer  \\ \midrule
    \multirow{2}{*}{$r = 3$}  & 0.8385 $\pm$ 0.0157 & 0.6144 $\pm$ 0.0094 & 0.8068 $\pm$ 0.0089 & \textbf{0.7457} $\pm$ \textbf{0.0078}  \\
     & (rel. err. 15.8\%) & (rel. err. 19.1\%) & (rel. err. 18.3\%) & (rel. err. 6.8\%)  \\[0.05in]
    \multirow{2}{*}{$r = 5$}  & \textbf{0.8527} $\pm$ \textbf{0.0129} & 0.6406 $\pm$ 0.0192 & \textbf{0.8205} $\pm$ \textbf{0.0093} & \textbf{0.7469} $\pm$ \textbf{0.0073} \\
     & (rel. err. 7.7\%) & (rel. err. 11.0\%) & (rel. err. 11.7\%) & (rel. err. 2.3\%) \\[0.05in]
    \multirow{2}{*}{$r = 10$} & \textbf{0.8465} $\pm$ \textbf{0.0103} & \textbf{0.6507} $\pm$ \textbf{0.0133} & \textbf{0.8259} $\pm$ \textbf{0.0132} & \textbf{0.7464} $\pm$ \textbf{0.0102} \\
     & (rel. err. 3.0\%) & (rel. err. 4.7\%) & (rel. err. 5.6\%) & (rel. err. 0.6\%) \\[0.05in]
    \multirow{2}{*}{$r = 20$} & \textbf{0.8540} $\pm$ \textbf{0.0078} & \textbf{0.6663} $\pm$ \textbf{0.0183} & \textbf{0.8272} $\pm$ \textbf{0.0159} & \textbf{0.7512} $\pm$ \textbf{0.0055} \\
     & (rel. err. 1.1\%) & (rel. err. 1.7\%) & (rel. err. 1.6\%) & (rel. err. 0.1\%) \\[0.05in]
    Exact computation & \textbf{0.8503} $\pm$ \textbf{0.0176} & \textbf{0.6506} $\pm$ \textbf{0.0199} & \textbf{0.8264} $\pm$ \textbf{0.0131} & \textbf{0.7481} $\pm$ \textbf{0.0076} \\ \bottomrule \\ \toprule
    & Coauthor CS & Coauthor Physics & Amazon Computer & Amazon Photo  \\ \midrule
    \multirow{2}{*}{$r = 3$}  & 0.8446 $\pm$ 0.0129                   & 0.9224 $\pm$ 0.0060                   & 0.7880 $\pm$ 0.0163                   & 0.8598 $\pm$ 0.0202 \\
     & (rel. err. 37.3\%) & (rel. err. 33.3\%) & (rel. err. 24.0\%) & (rel. err. 22.0\%) \\[0.05in]
    \multirow{2}{*}{$r = 5$}  & 0.8817 $\pm$ 0.0063                   & 0.9429 $\pm$ 0.0060                   & 0.8009 $\pm$ 0.0143                   & \textbf{0.8893} $\pm$ \textbf{0.0091} \\
     & (rel. err. 20.9\%) & (rel. err. 19.1\%) & (rel. err. 15.3\%) & (rel. err. 14.1\%) \\[0.05in]
    \multirow{2}{*}{$r = 10$} & \textbf{0.9018} $\pm$ \textbf{0.0115} & \textbf{0.9460} $\pm$ \textbf{0.0052} & \textbf{0.8265} $\pm$ \textbf{0.0197} & \textbf{0.9002} $\pm$ \textbf{0.0067} \\
     & (rel. err. 8.5\%) & (rel. err. 8.2\%) & (rel. err. 8.0\%) & (rel. err. 7.4\%) \\[0.05in]
    \multirow{2}{*}{$r = 20$} & \textbf{0.9018} $\pm$ \textbf{0.0077} & \textbf{0.9475} $\pm$ \textbf{0.0061} & \textbf{0.8342} $\pm$ \textbf{0.0143} & \textbf{0.9009} $\pm$ \textbf{0.0214} \\
    & (rel. err. 2.7\%) & (rel. err. 2.9\%) & (rel. err. 3.9\%) & (rel. err. 3.6\%) \\[0.05in]
    Exact computation & \textbf{0.9072} $\pm$ \textbf{0.0078} & \textbf{0.9502} $\pm$ \textbf{0.0051} & \textbf{0.8324} $\pm$ \textbf{0.0190} & \textbf{0.8971} $\pm$ \textbf{0.0121} \\ \bottomrule
    \end{tabular}
    \label{table: perf_experiments}
\end{table*}

Although \cite{GraphSAGE} already investigated the effect of node sampling to downstream tasks as a heuristic method, we also investigate it for completeness. We use eight node classification datasets, Cora, Cora Full, PubMed, Citeseer, Coauthor CS, Coauthor Physics, Amazon Computer, and Amazon Photo, retrieved from Deep Graph Library \url{https://docs.dgl.ai/api/python/data.html}. Cora, Cora Full, PubMed, and Citeseer are citation networks, Coauthor CS and Coauthor Physics are co-authorship networks, and Amazon Computer and Amazon Photo are co-purchase networks.
We train two-layered GraphSAGE-GCN with the neighbor sampling. We set the size of neighbor samples as $r_1 = r_2 = r$ both in training and inference. We use $500$ nodes for testing, $1000$ nodes for validation, and the remaining nodes for training for all datasets. Table \ref{table: perf_experiments} reports the average F1-score for $10$ different random seeds. We can see that even small numbers of neighbor samples offer good performance. In particular, $r = 3$ may be not enough, but $r = 5$ performs well in many datasets. Moreover, $r = 10$ and $r = 20$ perform as good as exact computation. These results validate the sampling approach in practical situations with various datasets. Table \ref{table: perf_experiments} also reports the relative approximation errors $\|\boldz_\text{exact} - \boldz_\text{approx}\| / \|\boldz_\text{exact}\|$ of the node embeddings in the last layer. It shows that (i) the approximation errors decrease as the number of sampled nodes increases, and (ii) most nodes are classified to the same category as in the exact computation when the relative error is around less than ten percent. Note that even if the embedding is not exact, the node is classified correctly unless the error crosses the classification boundary. This result shows that five to ten percent approximation suffices in typical downstream tasks.

\section{Practical Implications}

We summarize the practical implications of this work for practitioners.

\begin{itemize}
    \item We showed that GNNs can be approximated in constant time. When real time responses are required as the examples of the social network service and browser add-on we introduced in the introduction, we can use node sampling and estimate the approximation error using Theorem \ref{time}.
    \item We showed the lower bound of the approximation error in the worst case (Theorem \ref{opt}). We need at least $r = \Omega(\frac{1}{\varepsilon^2})$ samples to bound the approximation error in the worst case. We should keep this in mind when we apply an approximation technique, including other methods than node sampling, to applications that require guarantees.
    \item We showed that many of GNN variants cannot be approximated in constant time in Section \ref{sec:inapproximability}. We should be careful when we combine an approximation technique, including other methods than node sampling, with these GNN architectures. For example, when the performance of these GNN models is poor, we should investigate the sampling error in these cases.
    \item We showed that GCNs cannot be approximated in constant time when the degree distribution is skewed, but they can be approximated in constant time when the degree distribution is flat (Proposition \ref{gcn}, Assumption 6, experiments for Q5). This indicates that the approximation of GCNs requires close supervision when the input graph is a social network because the degree distribution of a social network presents the power law.
    \item Propositions \ref{gcn} and \ref{prop: GCN} also indicate that GCNs do use the degree distribution information in its inference, whereas other GNNs, such as GraphSAGE-GCN, do not exploit it. Therefore, GCNs are more appropriate than GraphSAGE-GCN when the degree distribution is important to improve performance. For example, when one wants to estimate the influence of users in a social network, they should use GCNs instead of GraphSAGE-GCN.
    \item We conducted numerical experiments using real world datasets (experiments for Q6). They describe the behavior of the node sampling in typical real world datasets. One can refer to them to estimate the approximation error in other real world datasets.
\end{itemize}

\section{Conclusion}
\label{conclusion}

We analyzed neighbor sampling to prove that it can approximate the embedding and gradient of GNNs in constant time, where the complexity is completely independent of the number of the nodes, edges, and neighbors of the input. This is the first analysis that offers constant time approximation for GNNs. We further demonstrated that some existing GNNs cannot be approximated in constant time by any algorithm. Finally, we validated the theory through experiments using synthetic and real-world datasets.

\section{Proofs} \label{sec: proof}

\begin{lemma}[Hoeffding's inequality \cite{Hoeffding}] \label{hoeffding}
Let $X_1, X_2, \dots, X_n$ be independent random variables bounded by the intervals $[-B, B]$ and let $\bar{X}$ be the empirical mean of these variables $\bar{X} = \frac{1}{n} \sum_{i = 1}^{n} X_i$. Then, for any $\varepsilon > 0$, 
\[ \textnormal{Pr}[|\bar{X} - \mathbb{E}[\bar{X}]| \ge \varepsilon] \le 2 \exp\left(-\frac{n\varepsilon^2}{2B^2}\right) \]
holds true.
\end{lemma}

\begin{lemma}[multivariate Hoeffding's inequality] \label{hoeffding_multi}
Let $\boldx_1, \boldx_2, \dots, \boldx_n$ be independent $d$-dimensional random variables whose two-norms are bounded $\|\boldx_i\|_2 \le B$, and let $\bar{\boldx}$ be the empirical mean of these variables $\bar{\boldx} = \frac{1}{n}\sum_{i = 1}^{n} \boldx_i$. Then, for any $\varepsilon > 0$, 
\[ \textnormal{Pr}[\|\bar{\boldx} - \mathbb{E}[\bar{\boldx}]\|_2 \ge \varepsilon] \le 2 d \exp\left(-\frac{n\varepsilon^2}{2B^2d}\right) \]
holds true.
\end{lemma}

\begin{proofln}{\ref{hoeffding_multi}}
Apply Lemma \ref{hoeffding} to each dimension $k$ of $X_i$. Then,
\[ \textnormal{Pr}[|\bar{X}_k - \mathbb{E}[\bar{X}]_k| \ge \frac{\varepsilon}{\sqrt{d}}] \le 2 \exp \left(- \frac{n \varepsilon^2}{2 B^2 d}\right). \]
It should be noted that $|X_{ik}| < B$ because $\|X_i\|_2 < B$. Therefore,
\begin{eqnarray*}
    && \textnormal{Pr}[\exists k \in \{1, 2, \dots, d\}~|\bar{X}_k - \mathbb{E}[\bar{X}]_k| \ge \frac{\varepsilon}{\sqrt{d}}] \le 2d \exp \left(- \frac{n \varepsilon^2}{2 B^2 d}\right).
\end{eqnarray*}
If $|\bar{X}_k - \mathbb{E}[\bar{X}]_k| < \frac{\varepsilon}{\sqrt{d}}$ holds true for all dimension $k$, then
\begin{align*}
\|\bar{X} - \mathbb{E}[\bar{X}]\|_2 = \sqrt{\sum_{k=1}^d (\bar{X}_k - \mathbb{E}[\bar{X}]_k)^2} 
< \sqrt{d \cdot \frac{\varepsilon^2}{d}} = \varepsilon.
\end{align*}
Therefore,
\[ \textnormal{Pr}[\|\bar{X} - \mathbb{E}[\bar{X}]\|_2 \ge \varepsilon] \le 2 d \exp\left(-\frac{n\varepsilon^2}{2B^2d}\right). \]
\end{proofln}

\begin{lemma} \label{zbound}
Under Assumptions 1 and 2, the norms of the embeddings $\|\boldz^{(l)}_v\|_2$, $\|\hat{\boldz}^{(l)}_v\|_2$, $\|\boldh^{(l)}_v\|_2$, and $\|\hat{\boldh}^{(l)}_v\|_2 ~(l = 1, \dots, L)$ are bounded by a constant $B \in \mathbb{R}$.
\end{lemma}

\begin{proofln}{\ref{zbound}}
We prove the theorem by performing mathematical induction. The norm of the input to the first layer is bounded by Assumption 1. The message function $\text{deg}(i) M_{liu}$ and the update function $U_{l}$ is continuous by Assumption 2. Since the image $f(X)$ of a compact set $X \in \mathbb{R}^{d}$ is compact if $f$ is continuous, the images of $\text{deg}(i) M_{liu}$ and $U_{l}$ are bounded by induction.
\end{proofln}

\begin{prooftn}{\ref{inference}}
We prove the theorem by performing mathematical induction on the number of layers $L$.

\noindent \textbf{Base case:} We show that the statement holds true for $L = 1$.
Because $U_{L}$ is uniform continuous,
\begin{align}
\exists \varepsilon' > 0, \forall \boldz_v, \boldh_v, \boldh'_v, \boldtheta, \| \boldh_v - \boldh'_v \|_2 < \varepsilon' \Rightarrow \|U_{L}(\boldz_v, \boldh_v, \boldtheta) - U_{L}(\boldz_v, \boldh'_v, \boldtheta)\|_2 < \varepsilon \label{eq: inference_Ubound_base}.
\end{align}
Let $x_k$ be the $k$-th sample in $\mathcal{S}^{(L)}$ and $X_k = \text{deg}(v) M_{Lvu}(\boldz^{(0)}_v, \boldz^{(0)}_{x_k}, \bolde_{vx_k}, \boldtheta)$. Then, 
\begin{align}
\mathbb{E} [X_k] = \sum_{u \in \mathcal{N}(v)} M_{Lvu}(\boldz^{(0)}_v, \boldz^{(0)}_u, \bolde_{vu}, \boldtheta) = \boldh^{(L)}_v \label{eq: inference_expectation_base}.
\end{align}
There exists a constant $C \in \mathbb{R}$ such that for any input satisfying Assumption 1, 
\begin{align}
\| X_k \|_2 < C \label{eq: inference_bound_base}
\end{align}
holds true because $\| \boldz^{(0)}_v \|_2, \| \boldz^{(0)}_{x_k} \|_2, \| \bolde_{vx_k} \|_2$, and $\| \boldtheta \|_2$ are bounded by Assumption 1 and $\text{deg}(v) M_{Lvu}$ is continuous. 
Therefore, if we take $r^{(L)} = O(\frac{1}{\varepsilon'^{2}} \log \frac{1}{\delta})$ samples, $\textnormal{Pr}[\|\hat{\boldh}_v^{(L)} - \boldh_v^{(L)}\|_2 \ge \varepsilon'] \le \delta$ by the Hoeffding's inequality and equations (\ref{eq: inference_expectation_base}) and (\ref{eq: inference_bound_base}). Therefore, $\textnormal{Pr}[\|\hat{\boldz}_v^{(L)} - \boldz_v^{(L)}\|_2 \ge \varepsilon] \le \delta$. We note that the input features are fixed in this analysis and the only source of randomness is sampling neighbor nodes. Thus $X_k$ is i.i.d. and we can apply the Hoeffding's inequality.

\noindent \textbf{Inductive step:} We show that the statement holds true for $L = l+1$ if it holds true for $L = l$. The induction hypothesis is $\forall \varepsilon > 0, 1 > \delta > 0$, $\exists r^{(1)}(\varepsilon, \delta), \dots, r^{(L-1)}(\varepsilon, \delta)$ such that $\forall v \in \mathcal{V}$, $\textnormal{Pr}[\|\hat{\mathcal{O}}_{z}^{(L-1)}(v, \varepsilon, \delta) - \boldz_v\|_2 \ge \varepsilon] \le \delta$.

Because $U_{L}$ is uniform continuous,

\begin{align}
\exists \varepsilon' > 0, \forall \boldz_v, \boldh_v, \boldz'_v, \boldh'_v, \boldtheta, \| [\boldz_v, \boldh_v] - [\boldz'_v, \boldh'_v] \|_2 < \varepsilon' \Rightarrow \|U_{L}(\boldz_v, \boldh_v, \boldtheta) - U_{L}(\boldz'_v, \boldh'_v, \boldtheta)\|_2 < \varepsilon \label{eq: inference_Ubound_induction},
\end{align}
where $[\cdot]$ denotes concatenation. By the induction hypothesis,
\begin{align}
\exists r'^{(1)}, \dots, r'^{(L-1)} \text{ such that } \textnormal{Pr}[\| \hat{\mathcal{O}}^{(L-1)}_z(v) - \boldz^{(L-1)}_v \|_2 \ge \varepsilon' / \sqrt{2}] \le \delta / 2 \label{eq: inference_U_hypothesis}.
\end{align}
holds true. Let 
\[ \tilde{\boldh}_{v}^{(L)} = \frac{\text{deg}(v)}{r^{(L)}} \sum_{u \in \mathcal{S}^{(L)}} M_{Lvu}(\boldz^{(L-1)}_v, \boldz^{(L-1)}_u, \bolde_{vu}, \boldtheta). \]
Let $x_k$ be the $k$-th sample in $\mathcal{S}^{(L)}$ and $X_k = \text{deg}(v) M_{Lvu}(\boldz_v^{(L-1)}, \boldz_{x_k}^{(L-1)}, \bolde_{vx_k}, \boldtheta)$. Then, 
\begin{align}
\mathbb{E} [X_k] = \sum_{u \in \mathcal{N}(v)} M_{Lvu}(\boldz_v^{(L-1)}, \boldz_{x_k}^{(L-1)}, \bolde_{vx_k}, \boldtheta) = \boldh^{(L)}_v \label{eq: inference_expectation_induction}.
\end{align}
There exists a constant $C \in \mathbb{R}$ such that for any input satisfying Assumption 1, 
\begin{align}
\| X_k \|_2 < C \label{eq: inference_bound_induction},
\end{align}
because $\| \boldz^{(L-1)}_v \|_2, \| \boldz^{(L-1)}_{x_k} \|_2, \| \bolde_{vx_k} \|_2$, and $\| \boldtheta \|_2$ are bounded by Assumption 1 and Theorem \ref{zbound}, and $\text{deg}(v) M_{Lvu}$ is continuous. 
If we take $r^{(L)} = O(\frac{1}{\varepsilon'^{2}} \log \frac{1}{\delta})$, then 
\begin{align}
\textnormal{Pr}[\| \tilde{\boldh}^{(l)}_v - \boldh^{(l)}_v \|_2 \ge \varepsilon' / (2 \sqrt{2})] \le \delta / 4 \label{eq: inference_tri1}, 
\end{align}
by the Hoeffding's inequality and equations (\ref{eq: inference_expectation_induction}) and (\ref{eq: inference_bound_induction}).
Because $\text{deg}(v) M_{Lvu}$ is uniform continuous,
\begin{align}
& \exists \varepsilon'' > 0 \text{ such that } \| [\boldz^{(L-1)}_v, \boldz^{(L-1)}_u] - [\boldz'^{(L-1)}_v, \boldz'^{(L-1)}_u] \|_2 \le \varepsilon'' \notag \\
& \quad \Rightarrow \text{deg}(v) \| M_{Lvu}(\boldz^{(L-1)}_v, \boldz^{(L-1)}_u, \bolde_{vu}, \boldtheta) - M_{Lvu}(\boldz'^{(L-1)}_v, \boldz'^{(L-1)}_u, \bolde_{vu}, \boldtheta) \|_2 \le \varepsilon' / (2 \sqrt{2}) \label{eq: inference_Mbound} .
\end{align}
By the induction hypothesis,
\begin{align}
\exists r''^{(1)}, \dots, r''^{(l)} \text{ such that } \textnormal{Pr}[\| \hat{\mathcal{O}}^{(L-1)}_z(v) - \boldz^{(L-1)}_v \|_2 \ge \varepsilon'' / \sqrt{2}] \le \delta / (8 r^{(L)}) \label{inference_M_hypothesis}.
\end{align}
Therefore, the probability that the errors of all oracle calls are bounded is
\begin{align}
\textnormal{Pr}[\exists u \in \mathcal{S}^{(L)}, \| [\hat{\mathcal{O}}^{(L-1)}_z(v), \hat{\mathcal{O}}^{(L-1)}_z(u)] - [\boldz^{(L-1)}_v, \boldz^{(L-1)}_u] \|_2 \ge \varepsilon''] \le \delta / 4 \label{eq: inference_induction_call}.
\end{align}
By equations (\ref{eq: inference_Mbound}) and (\ref{eq: inference_induction_call}),
\begin{align*}
\textnormal{Pr}[&\exists u \in \mathcal{S}^{(L)}, \\
&\text{deg}(v) \| M_{Lvu}(\boldz^{(L-1)}_v, \boldz^{(L-1)}_u, \bolde_{vu}, \boldtheta) - M_{Lvu}(\hat{\boldz}^{(L-1)}_v, \hat{\boldz}^{(L-1)}_u, \bolde_{vu}, \boldtheta) \|_2 \ge \varepsilon' / (2 \sqrt{2})] \le \delta / 4 .
\end{align*}
\begin{align}
\textnormal{Pr}[\| \tilde{\boldh}^{(L)}_v - \hat{\boldh}^{(L)}_v \|_2 \ge \varepsilon' / (2 \sqrt{2})] \le \delta / 4 \label{eq: inference_tri2}. 
\end{align}
By the triangular inequality and equations (\ref{eq: inference_tri1}) and (\ref{eq: inference_tri2}),
\begin{align}
\textnormal{Pr}[\| \hat{\boldh}^{(L)}_v - \boldh^{(L)}_v \|_2 \ge \varepsilon' / \sqrt{2}] \le \delta / 2 \label{eq: inference_U_h}.
\end{align}
Therefore, if we take $r^{(1)} = \max(r'^{(1)}, r''^{(1)}), \dots, r^{(L-1)} = \max(r'^{(L-1)}, r''^{(L-1)})$, by equations (\ref{eq: inference_U_hypothesis}) and (\ref{eq: inference_U_h}),
\begin{align}
\textnormal{Pr}[ \| [\boldz_v^{(L-1)}, \boldh_v^{(L)}] - [\hat{\mathcal{O}}^{(L-1)}_z(v), \hat{\boldh}_v^{(L)}] \|_2 \ge \varepsilon'] \le \delta \label{eq: inference_induction_h}.
\end{align}
Therefore, by equations (\ref{eq: inference_Mbound}) and (\ref{eq: inference_induction_h}),
\[\textnormal{Pr}[\| \hat{\boldz}^{(L)}_v - \boldz^{(L)}_v \|_2 \ge \varepsilon] \le \delta. \]

\end{prooftn}

\begin{prooftn}{\ref{time}}
We prove this by performing mathematical induction on the number of layers.

\noindent \textbf{Base case:} We show that the statement holds true for $L = 1$.
If $U_L$ is $K$-Lipschitz continuous, $\varepsilon' = O(\varepsilon)$ in equation (\ref{eq: inference_Ubound_base}). Therefore, $r^{(L)} = O(\frac{1}{\varepsilon^{2}} \log \frac{1}{\delta})$.

\noindent \textbf{Inductive step:} We show that the statement holds true for $L = l+1$ if it holds true for $L = l$.
If $U_L$ and $M_{Lvu}$ are $K$-Lipschitz continuous, $\varepsilon' = O(\varepsilon)$ in equation (\ref{eq: inference_Ubound_induction}) and $\varepsilon'' = O(\varepsilon)$ in equation (\ref{eq: inference_Mbound}). Therefore, $r^{(L)} = O(\frac{1}{\varepsilon^{2}} \log \frac{1}{\delta})$. 
We call $\hat{\mathcal{O}}^{(L-1)}_z(v)$ such that $\textnormal{Pr}[\| \hat{\mathcal{O}}^{(L-1)}_z(v) - \boldz^{(L-1)}_v \|_2 \ge \varepsilon' / \sqrt{2}] \le \delta / 2$ in equation (\ref{eq: inference_U_hypothesis}). Therefore, $r'^{(1)}, \dots, r'^{(L-1)} = O(\frac{1}{\varepsilon^2} (\log \frac{1}{\varepsilon} + \log \frac{1}{\delta}))$ are sufficient by the induction hypothesis.
We call $\hat{\mathcal{O}}^{(L-1)}_z(v)$ such that $\textnormal{Pr}[\| \hat{\mathcal{O}}^{(L-1)}_z(v) - \boldz^{(L-1)}_v \|_2 \ge \varepsilon'' / \sqrt{2}] \le \delta / (8 r^{(L)})$ in equation (\ref{inference_M_hypothesis}). Therefore, $r''^{(1)}, \dots, r''^{(L-1)} = O(\frac{1}{\varepsilon^2} (\log \frac{1}{\varepsilon} + \log \frac{1}{\delta}))$ are sufficient by the induction hypothesis because $\log \frac{1}{\delta / (8 r^{(L)})} = O(\log \frac{1}{\varepsilon} + \log \frac{1}{\delta})$. In total, the complexity is $O(\frac{1}{\varepsilon^{2L}} (\log \frac{1}{\varepsilon} + \log \frac{1}{\varepsilon})^{L-1} \log \frac{1}{\delta})$
\end{prooftn}

\begin{lemma} [\cite{chazelle2005approximating}] \label{bp}
Let $\mathcal{D}^s$ be $Bernoulli(\frac{1 + s \varepsilon}{2})$. Let $n$-dimentional distribution $\mathcal{D}$ be (1) pick $s = 1$ with probability $1/2$ and $s = -1$ otherwise; (2) then draw $n$ values from $\mathcal{D}^s$. Any probabilistic algorithm that can guess the value of $s$ with a probability error below $1/4$ requires $\Omega(\frac{1}{\varepsilon^2})$ bit lookup on average.
\end{lemma}

\begin{prooftn}{\ref{opt}}
We prove there is a counter example in the GraphSAGE-GCN models. Suppose there is an algorithm that approximates the one-layer GraphSAGE-GCN within $o(\varepsilon^2)$ queries. We prove that this algorithm can distinguish $\mathcal{D}$ in Lemma \ref{bp} within $o(\varepsilon^2)$ queries and derive a contradiction. 

Let $\sigma$ be any non-constant $K$-Lipschitz activation function. There exists $a, b \in \mathbb{R} ~(a > b)$ such that $\sigma(a) \neq \sigma(b)$ because $\sigma$ is not constant. Let $S = \frac{|\sigma(a) - \sigma(b)|}{a - b} > 0$. Let $\varepsilon > 0$ be any sufficiently small positive value and $t \in \{0, 1\}^n$ be a random variable drawn from $\mathcal{D}$. We prove that we can determine $s$ with high provability within $o(\varepsilon^2)$ queries using the algorithm. Let $G$ be a clique $K_n$ and $\boldW^{(1)} = 1$. Let us calculate $a_{\varepsilon}$ and $b_{\varepsilon}$ using the following steps: (1) set $a_{\varepsilon} = a$ and $b_{\varepsilon} = b$; (2) if $a_{\varepsilon} - b_{\varepsilon} < \varepsilon$, return $a_{\varepsilon}$ and $b_{\varepsilon}$; (3) $m = \frac{a_{\varepsilon} + b_{\varepsilon}}{2}$; (4) if $|\sigma(a_{\varepsilon}) - \sigma(m)| > |\sigma(m) - \sigma(b_{\varepsilon})|$, then set $b_{\varepsilon} = m$, otherwise $a_{\varepsilon} = m$; and (5) go back to (2). Here, $\varepsilon / 2 \le a_{\varepsilon} - b_{\varepsilon} < \varepsilon$, $a \le \frac{a_{\varepsilon} + b_{\varepsilon}}{2} \le b$, and $|\sigma(a_{\varepsilon}) - \sigma(b_{\varepsilon})| \ge \frac{S}{2} \varepsilon$ hold true. Let $x_v = \frac{a_{\varepsilon} + b_{\varepsilon}}{2} + (2t_v - 1) \frac{a_{\varepsilon} - b_{\varepsilon}}{2 \varepsilon}$ for all $v \in \mathcal{V}$. Then, $\mathbb{E}[h_v \mid s = 1] = a_{\varepsilon}$ and $\mathbb{E}[h_v \mid s = -1] = b_{\varepsilon}$. Therefore, $\textnormal{Pr}[|z_v - \sigma(a_{\varepsilon})| < \frac{S}{8} \varepsilon \mid s = 1] \to 1$ as $n \to \infty$ and $\textnormal{Pr}[|z_v - \sigma(b_{\varepsilon})| < \frac{S}{8} \varepsilon \mid s = -1] \to 1$ as $n \to \infty$ because $\sigma$ is $K$-Lipschitz. We set the error tolerance to $\frac{S}{8} \varepsilon$ and $n$ to a sufficiently large number. Then $s = 1$ if $|\hat{z}_v - \sigma(a_{\varepsilon})| < \frac{S}{4} \varepsilon$ and $s = -1$ otherwise with high probability. However, the algorithm accesses $t$ (i.e., accesses $\mathcal{O}_{\text{feature}}$) $o(\varepsilon^2)$ times. This contradicts with Lemma \ref{bp}.
\end{prooftn}

\begin{lemma} \label{gbound}
Under Assumptions 1, 2 and 3, the norms of the gradients of the message functions and the update functions $\| D U_l(\boldz^{(l-1)}_v, \boldh^{(l)}_v, \boldtheta) \|_F$, $\| D U_l(\hat{\boldz}^{(l-1)}_v, \hat{\boldh}^{(l)}_v, \boldtheta) \|_F$, $\| \text{deg}(v) D M_{lvu}(\boldz^{(l-1)}_v, \boldv^{(l)}_u, \bolde_{vu}, \boldtheta) \|_F$, and $\| \text{deg}(v) D M_{lvu}(\hat{\boldz}^{(l-1)}_v, \hat{\boldv}^{(l)}_u, \bolde_{vu}, \boldtheta) \|_F$ are bounded by a constant $B' \in \mathbb{R}$.
\end{lemma}

\begin{proofln}{\ref{gbound}}
The input of each function is bounded by Lemma \ref{zbound}. Because $D U_l$ and $\text{deg}(v) D M_{lvu}$ is uniform continuous, these images are bounded.
\end{proofln}

\begin{prooftn}{\ref{gradient}}
We prove the theorem by performing mathematical induction on the number of layers $L$.

\noindent \textbf{Base case:} We show that the statement holds true for $L = 1$.

When the number of layers is one,

\begin{align*}
\frac{\partial \boldz^{(L)}_v}{\partial \boldtheta} &= \frac{\partial U_{L}}{\partial \boldtheta} (\boldz^{(0)}_v, \boldh^{(L)}_v, \boldtheta) + \frac{\partial U_{L}}{\partial \boldh^{(L)}_v} (\boldz^{(0)}_v, \boldh^{(L)}_v, \boldtheta) \frac{\partial \boldh^{(L)}_v}{\partial \boldtheta} \\
&= \frac{\partial U_{L}}{\partial \boldtheta} (\boldz^{(0)}_v, \boldh^{(L)}_v, \boldtheta) + \frac{\partial U_{L}}{\partial \boldh^{(L)}_v} (\boldz^{(0)}_v, \boldh^{(L)}_v, \boldtheta) \sum_{u \in \mathcal{N}(v)} \frac{\partial M_{Lvu}}{\partial \boldtheta} (\boldz^{(0)}_v, \boldz^{(0)}_u, \bolde_{vu}, \boldtheta).
\end{align*}

\begin{align*}
\frac{\partial \hat{\boldz}^{(L)}_v}{\partial \boldtheta} &= \frac{\partial U_{L}}{\partial \boldtheta} (\boldz^{(0)}_v, \hat{\boldh}^{(L)}_v, \boldtheta) + \frac{\partial U_{L}}{\partial \hat{\boldh}^{(L)}_v} (\boldz^{(0)}_v, \hat{\boldh}^{(L)}_v, \boldtheta) \frac{\partial \hat{\boldh}^{(L)}_v}{\partial \boldtheta} \\
&= \frac{\partial U_{L}}{\partial \boldtheta} (\boldz^{(0)}_v, \hat{\boldh}^{(L)}_v, \boldtheta) + \frac{\partial U_{L}}{\partial \hat{\boldh}^{(L)}_v} (\boldz^{(0)}_v, \hat{\boldh}^{(L)}_v, \boldtheta) \frac{\text{deg}(v)}{r^{(L)}} \sum_{u \in \mathcal{S}^{(L)}} \frac{\partial M_{Lvu}}{\partial \boldtheta} (\boldz^{(0)}_v, \boldz^{(0)}_u, \bolde_{vu}, \boldtheta).
\end{align*}

Because $D U_{L}$ is uniform continuous,

\begin{align}
    &\exists \varepsilon' > 0 \text{ s.t. for all input,} \| \boldh^{(L)}_v - \hat{\boldh}^{(L)} \|_2 < \varepsilon' \Rightarrow \notag \\
    &(\| \frac{\partial U_{L}}{\partial \boldtheta} (\boldz^{(0)}_v, \boldh^{(L)}_v, \boldtheta) - \frac{\partial U_{L}}{\partial \boldtheta} (\boldz^{(0)}_v, \hat{\boldh}^{(L)}_v, \boldtheta) \|_F < \varepsilon / 2 ~\land \notag \\
    &\| \frac{\partial U_{L}}{\partial \boldh^{(L)}_v} (\boldz^{(0)}_v, \boldh^{(L)}_v, \boldtheta) -  \frac{\partial U_{L}}{\partial \hat{\boldh}^{(L)}_v} (\boldz^{(0)}_v, \hat{\boldh}^{(L)}_v, \boldtheta) \|_F < \varepsilon / (4B')) \label{eq: grad_Ubound_base}.
\end{align}
If we take $r^{(L)} = O(\frac{1}{\varepsilon'^2} \log \frac{1}{\delta})$,
\begin{align}
\textnormal{Pr}[ \| \boldh^{(L)}_v - \hat{\boldh}^{(L)} \|_2 \ge \varepsilon'] \le \delta / 2 \label{eq: grad_h_base}
\end{align}
holds true for any input by the argument of the proof of Theorem \ref{inference}.
Let $x_k$ be the $k$-th sample in $\mathcal{S}^{(L)}$ and $X_k = \text{deg}(v) \frac{\partial M_{Lvu}}{\partial \boldtheta} (\boldz^{(0)}_v, \boldz^{(0)}_u, \bolde_{vu}, \boldtheta)$. Then, 
\begin{align}
\mathbb{E} [X_k] = \sum_{u \in \mathcal{N}(v)} \frac{\partial M_{Lvu}}{\partial \boldtheta} (\boldz^{(0)}_v, \boldz^{(0)}_u, \bolde_{vu}, \boldtheta) = \frac{\partial \boldh^{(L)}_v}{\partial \boldtheta} \label{eq: gradient_expectation_base}.
\end{align}
There exists a constant $C \in \mathbb{R}$ such that for any input satisfying Assumption 1, 
\begin{align}
\| X_k \|_2 < C \label{eq: gradient_bound_base},
\end{align}
because $\| \boldz^{(0)}_v \|_2, \| \boldz^{(0)}_{x_k} \|_2, \| \bolde_{vx_k} \|_2$, and $\| \boldtheta \|_2$ are bounded by Assumption 1 and $\text{deg}(v) D M_{Lvu}$ is continuous. 
Therefore, if we take $r^{(L)} = O(\frac{1}{\varepsilon^{2}} \log \frac{1}{\delta})$ samples, 
\begin{align}
\textnormal{Pr}[\|\frac{\partial \hat{\boldh}_v^{(L)}}{\partial \boldtheta} - \frac{\partial \boldh_v^{(L)}}{\partial \boldtheta} \|_F \ge \varepsilon / (4B')] \le \delta / 2 \label{eq: grad_h_grad_base}
\end{align}
holds true by the Hoeffding's inequality and equations (\ref{eq: gradient_expectation_base}) and (\ref{eq: gradient_bound_base}).  If 
\[ \|\frac{\partial \hat{\boldh}_v^{(L)}}{\partial \boldtheta} - \frac{\partial \boldh_v^{(L)}}{\partial \boldtheta} \|_F < \varepsilon / (4B') \]
and
\[ \| \frac{\partial U_{L}}{\partial \boldh^{(L)}_v} (\boldz^{(0)}_v, \boldh^{(L)}_v, \boldtheta) -  \frac{\partial U_{L}}{\partial \hat{\boldh}^{(L)}_v} (\boldz^{(0)}_v, \hat{\boldh}^{(L)}_v, \boldtheta) \|_F < \varepsilon / (4B')) \]
hold true, then
\begin{align*}
& \| \frac{\partial \hat{\boldh}_v^{(L)}}{\partial \boldtheta} \frac{\partial U_{L}}{\partial \hat{\boldh}^{(L)}_v} (\boldz^{(0)}_v, \hat{\boldh}^{(L)}_v, \boldtheta) - \frac{\partial \boldh_v^{(L)}}{\partial \boldtheta}  \frac{\partial U_{L}}{\partial \boldh^{(L)}_v} (\boldz^{(0)}_v, \boldh^{(L)}_v, \boldtheta) \|_F \\
&\le \| \frac{\partial \hat{\boldh}_v^{(L)}}{\partial \boldtheta} \frac{\partial U_{L}}{\partial \hat{\boldh}^{(L)}_v} (\boldz^{(0)}_v, \hat{\boldh}^{(L)}_v, \boldtheta) - \frac{\partial \hat{\boldh}_v^{(L)}}{\partial \boldtheta} \frac{\partial U_{L}}{\partial \boldh^{(L)}_v} (\boldz^{(0)}_v, \boldh^{(L)}_v, \boldtheta) \|_F \\
&\quad + \| \frac{\partial \hat{\boldh}_v^{(L)}}{\partial \boldtheta} \frac{\partial U_{L}}{\partial \boldh^{(L)}_v} (\boldz^{(0)}_v, \boldh^{(L)}_v, \boldtheta) - \frac{\partial \boldh_v^{(L)}}{\partial \boldtheta}  \frac{\partial U_{L}}{\partial \boldh^{(L)}_v} (\boldz^{(0)}_v, \boldh^{(L)}_v, \boldtheta) \|_F \\
&= \| \frac{\partial \hat{\boldh}_v^{(L)}}{\partial \boldtheta} \|_F \| \frac{\partial U_{L}}{\partial \hat{\boldh}^{(L)}_v} (\boldz^{(0)}_v, \hat{\boldh}^{(L)}_v, \boldtheta) - \frac{\partial U_{L}}{\partial \boldh^{(L)}_v} (\boldz^{(0)}_v, \boldh^{(L)}_v, \boldtheta) \|_F \\
&\quad + \| \frac{\partial \hat{\boldh}_v^{(L)}}{\partial \boldtheta} - \frac{\partial \boldh_v^{(L)}}{\partial \boldtheta} \|_F \| \frac{\partial U_{L}}{\partial \boldh^{(L)}_v} (\boldz^{(0)}_v, \boldh^{(L)}_v, \boldtheta) \|_F \\
&\le B' \| \frac{\partial U_{L}}{\partial \hat{\boldh}^{(L)}_v} (\boldz^{(0)}_v, \hat{\boldh}^{(L)}_v, \boldtheta) - \frac{\partial U_{L}}{\partial \boldh^{(L)}_v} (\boldz^{(0)}_v, \boldh^{(L)}_v, \boldtheta) \|_F + B' \| \frac{\partial \hat{\boldh}_v^{(L)}}{\partial \boldtheta} - \frac{\partial \boldh_v^{(L)}}{\partial \boldtheta} \|_F \\
&< B'\frac{\varepsilon}{4B'} + B'\frac{\varepsilon}{4B'} = \frac{\varepsilon}{2}.
\end{align*}

Therefore, $\textnormal{Pr}[\| \frac{\partial \boldz^{(L)}_v}{\partial \boldtheta} - \frac{\partial \boldz^{(L)}_v}{\partial \boldtheta}\|_F \ge \varepsilon] \le \delta$ by equations (\ref{eq: grad_Ubound_base}), (\ref{eq: grad_h_base}), and (\ref{eq: grad_h_grad_base}).

\noindent \textbf{Inductive step:} We show that the statement holds true for $L = l+1$ if it holds true for $L = l$.

\begin{align*}
\frac{\partial \boldz^{(L)}_v}{\partial \boldtheta} &= \frac{\partial U_{L}}{\partial \boldtheta} (\boldz^{(L-1)}_v, \boldh^{(L)}_v, \boldtheta) + \frac{\partial U_{L}}{\partial \boldz^{(L-1)}_v} (\boldz^{(L-1)}_v, \boldh^{(L)}_v, \boldtheta) \frac{\partial \boldz^{(L-1)}_v}{\partial \boldtheta} \\
&+ \frac{\partial U_{L}}{\partial \boldh^{(L)}_v} (\boldz^{(L-1)}_v, \boldh^{(L)}_v, \boldtheta) \sum_{u \in \mathcal{N}(v)} \frac{\partial M_{Lvu}}{\partial \boldtheta} (\boldz^{(L-1)}_v, \boldz^{(L-1)}_u, \bolde_{vu}, \boldtheta) \\
&+ \frac{\partial U_{L}}{\partial \boldh^{(L)}_v}(\boldz^{(L-1)}_v, \boldh^{(L)}_v, \boldtheta) \sum_{u \in \mathcal{N}(v)} \frac{\partial M_{Lvu}}{\partial \boldz^{(L-1)}_v}(\boldz^{(L-1)}_v, \boldz^{(L-1)}_u, \bolde_{vu}, \boldtheta) \frac{\partial \boldz^{(L-1)}_v}{\partial \boldtheta} \\
&+ \frac{\partial U_{L}}{\partial \boldh^{(L)}_v}(\boldz^{(L-1)}_v, \boldh^{(L)}_v, \boldtheta) \sum_{u \in \mathcal{N}(v)} \frac{\partial M_{Lvu}}{\partial \boldz^{(L-1)}_u}(\boldz^{(L-1)}_v, \boldz^{(L-1)}_u, \bolde_{vu}, \boldtheta) \frac{\partial \boldz^{(L-1)}_u}{\partial \boldtheta}.
\end{align*}

\begin{align*}
\frac{\partial \hat{\boldz}^{(L)}_v}{\partial \boldtheta} &= \frac{\partial U_{L}}{\partial \boldtheta} (\hat{\boldz}^{(L-1)}_v, \hat{\boldh}^{(L)}_v, \boldtheta) + \frac{\partial U_{L}}{\partial \hat{\boldz}^{(L-1)}_v} (\hat{\boldz}^{(L-1)}_v, \hat{\boldh}^{(L)}_v, \boldtheta) \frac{\partial \hat{\boldz}^{(L-1)}_v}{\partial \boldtheta} \\
&+ \frac{\partial U_{L}}{\partial \hat{\boldh}^{(L)}_v} (\hat{\boldz}^{(L-1)}_v, \hat{\boldh}^{(L)}_v, \boldtheta) \frac{\text{deg}(v)}{r^{(L)}} \sum_{u \in \mathcal{S}^{(L)}} \frac{\partial M_{Lvu}}{\partial \boldtheta} (\hat{\boldz}^{(L-1)}_v, \hat{\boldz}^{(L-1)}_u, \bolde_{vu}, \boldtheta) \\
&+ \frac{\partial U_{L}}{\partial \hat{\boldh}^{(L)}_v}(\hat{\boldz}^{(L-1)}_v, \hat{\boldh}^{(L)}_v, \boldtheta) \frac{\text{deg}(v)}{r^{(L)}} \sum_{u \in \mathcal{S}^{(L)}} \frac{\partial M_{Lvu}}{\partial \hat{\boldz}^{(L-1)}_v}(\hat{\boldz}^{(L-1)}_v, \hat{\boldz}^{(L-1)}_u, \bolde_{vu}, \boldtheta) \frac{\partial \hat{\boldz}^{(L-1)}_v}{\partial \boldtheta} \\
&+ \frac{\partial U_{L}}{\partial \hat{\boldh}^{(L)}_v}(\hat{\boldz}^{(L-1)}_v, \hat{\boldh}^{(L)}_v, \boldtheta) \frac{\text{deg}(v)}{r^{(L)}} \sum_{u \in \mathcal{S}^{(L)}} \frac{\partial M_{Lvu}}{\partial \hat{\boldz}^{(L-1)}_u}(\hat{\boldz}^{(L-1)}_v, \hat{\boldz}^{(L-1)}_u, \bolde_{vu}, \boldtheta) \frac{\partial \hat{\boldz}^{(L-1)}_u}{\partial \boldtheta}.
\end{align*}

Because $D U_{L}$ is uniform continuous,

\begin{align}
    &\exists \varepsilon' > 0 \text{ s.t. for all input,} \| [\boldz^{(L-1)}_v, \boldh^{(L)}_v] - [\hat{\boldz}^{(L-1)}_v, \hat{\boldh}^{(L)}_v] \|_2 < \varepsilon' \Rightarrow \notag \\
    &(\| \frac{\partial U_{L}}{\partial \boldtheta} (\boldz^{(L-1)}_v, \boldh^{(L)}_v, \boldtheta) - \frac{\partial U_{L}}{\partial \boldtheta} (\hat{\boldz}^{(L-1)}_v, \hat{\boldh}^{(L)}_v, \boldtheta) \|_F < O(\varepsilon) ~\land \notag \\
    &\| \frac{\partial U_{L}}{\partial \boldh^{(L)}_v} (\boldz^{(L-1)}_v, \boldh^{(L)}_v, \boldtheta) -  \frac{\partial U_{L}}{\partial \hat{\boldh}^{(L)}_v} (\hat{\boldz}^{(L-1)}_v, \hat{\boldh}^{(L)}_v, \boldtheta) \|_F < O(\varepsilon) ~\land \notag \\
    &\| \frac{\partial U_{L}}{\partial \boldz^{(L-1)}_v} (\boldz^{(L-1)}_v, \boldh^{(L)}_v, \boldtheta) -  \frac{\partial U_{L}}{\partial \hat{\boldz}^{(L-1)}_v} (\hat{\boldz}^{(L-1)}_v, \hat{\boldh}^{(L)}_v, \boldtheta) \|_F < O(\varepsilon)) \label{eq: grad_Ubound_induction}.
\end{align}
Here, $X < O(\varepsilon)$ means that there exists a universal constant $\alpha$ such that $X < \alpha \varepsilon$. Because $\text{deg}(v) D M_{Lvu}$ is uniform continuous,

\begin{align}
    &\exists \varepsilon'' > 0 \text{ s.t. for all input,} \| [\boldz^{(L-1)}_v, \boldz^{(L-1)}_u] - [\hat{\boldz}^{(L-1)}_v, \hat{\boldz}^{(L-1)}_u] \|_2 < \varepsilon'' \Rightarrow \notag \\
    &(\text{deg}(v) \| \frac{\partial M_{Lvu}}{\partial \boldtheta} (\boldz^{(L-1)}_v, \boldz^{(L-1)}_u, \bolde_{vu}, \boldtheta) - \frac{\partial U_{L}}{\partial \boldtheta} (\hat{\boldz}^{(L-1)}_v, \hat{\boldz}^{(L-1)}_u, \bolde_{vu}, \boldtheta) \|_F < O(\varepsilon) ~\land \notag \\
    &\text{deg}(v) \| \frac{\partial M_{Lvu}}{\partial \boldh^{(L)}_v} (\boldz^{(L-1)}_v, \boldz^{(L-1)}_u, \bolde_{vu}, \boldtheta) -  \frac{\partial U_{L}}{\partial \hat{\boldh}^{(L)}_v} (\hat{\boldz}^{(L-1)}_v, \hat{\boldz}^{(L-1)}_u, \bolde_{vu}, \boldtheta) \|_F < O(\varepsilon) ~\land \notag \\
    &\text{deg}(v) \| \frac{\partial M_{Lvu}}{\partial \boldz^{(L-1)}_v} (\boldz^{(L-1)}_v, \boldz^{(L-1)}_u, \bolde_{vu}, \boldtheta) -  \frac{\partial U_{L}}{\partial \hat{\boldz}^{(L-1)}_v} (\hat{\boldz}^{(L-1)}_v, \hat{\boldz}^{(L-1)}_u, \bolde_{vu}, \boldtheta) \|_F < O(\varepsilon)) \label{eq: grad_Mbound_induction}.
\end{align}
By the argument of the proof of Theorem \ref{inference}, if we take sufficiently large number of samples,
\begin{align}
\textnormal{Pr}[ \|\boldz^{(L-1)}_v - \hat{\boldz}^{(L-1)}_v\|_2 \ge O(\min(\varepsilon', \varepsilon''))] \le O(\varepsilon \delta), \label{eq: grad_z_induction}
\end{align}
\begin{align}
\textnormal{Pr}[ \|\boldh^{(L)}_v - \hat{\boldh}^{(L)}_v\|_2 \ge O(\varepsilon')] \le O(\delta) \label{eq: grad_h_induction}.
\end{align}
By the induction hypothesis, there exists $r^{(1)}, \dots, r^{(L-1)}$ such that 
\[ \textnormal{Pr}[ \| \frac{\partial \boldz^{(L-1)}_u}{\partial \boldtheta} - \frac{\partial \hat{\boldz}^{(L-1)}_u}{\partial \boldtheta} \|_F \ge O(\varepsilon)] \le O(\varepsilon \delta), \]
\[ \textnormal{Pr}[ \| \frac{\partial \boldz^{(L-1)}_v}{\partial \boldtheta} - \frac{\partial \hat{\boldz}^{(L-1)}_v}{\partial \boldtheta} \|_F \ge O(\varepsilon)] \le O(\varepsilon \delta), \]
If we take $r^{(L)} = O(\frac{1}{\varepsilon^2} \log \frac{1}{\delta})$,
\begin{align}
    \textnormal{Pr}[ \| &\sum_{u \in \mathcal{N}(v)} \frac{\partial M_{Lvu}}{\partial \boldtheta} (\boldz^{(L-1)}_v, \boldz^{(L-1)}_u, \bolde_{vu}, \boldtheta) \notag \\ &- \frac{\text{deg}(v)}{r^{(L)}} \sum_{u \in \mathcal{S}^{(L)}} \frac{\partial M_{Lvu}}{\partial \boldtheta} (\boldz^{(L-1)}_v, \boldz^{(L-1)}_u, \bolde_{vu}, \boldtheta) \| \ge O(\varepsilon)] \le O(\delta), \label{eq: gradient_hoeffding_1}
\end{align}
\begin{align}
    \textnormal{Pr}[ \| &\sum_{u \in \mathcal{N}(v)} \frac{\partial M_{Lvu}}{\partial \boldz^{(L-1)}_v}(\boldz^{(L-1)}_v, \boldz^{(L-1)}_u, \bolde_{vu}, \boldtheta) \frac{\partial \boldz^{(L-1)}_v}{\partial \boldtheta} \notag \\ &- \frac{\text{deg}(v)}{r^{(L)}} \sum_{u \in \mathcal{S}^{(L)}} \frac{\partial M_{Lvu}}{\partial \boldz^{(L-1)}_v}(\boldz^{(L-1)}_v, \boldz^{(L-1)}_u, \bolde_{vu}, \boldtheta) \frac{\partial \boldz^{(L-1)}_v}{\partial \boldtheta} \| \ge O(\varepsilon)] \le O(\delta), \label{eq: gradient_hoeffding_2}
\end{align}
\begin{align}
    \textnormal{Pr}[ \| &\sum_{u \in \mathcal{N}(v)} \frac{\partial M_{Lvu}}{\partial \boldz^{(L-1)}_u}(\boldz^{(L-1)}_v, \boldz^{(L-1)}_u, \bolde_{vu}, \boldtheta) \frac{\partial \boldz^{(L-1)}_u}{\partial \boldtheta} \notag \\ &- \frac{\text{deg}(v)}{r^{(L)}} \sum_{u \in \mathcal{S}^{(L)}} \frac{\partial M_{Lvu}}{\partial \boldz^{(L-1)}_u}(\boldz^{(L-1)}_v, \boldz^{(L-1)}_u, \bolde_{vu}, \boldtheta) \frac{\partial \boldz^{(L-1)}_u}{\partial \boldtheta} \| \ge O(\varepsilon)] \le O(\delta), \label{eq: gradient_hoeffding_3}
\end{align}
holds true by the Hoeffding's inequality. Therefore,
\begin{align}
    \textnormal{Pr}[ \| &\sum_{u \in \mathcal{N}(v)} \frac{\partial M_{Lvu}}{\partial \boldtheta} (\boldz^{(L-1)}_v, \boldz^{(L-1)}_u, \bolde_{vu}, \boldtheta) \notag \\ &- \frac{\text{deg}(v)}{r^{(L)}} \sum_{u \in \mathcal{S}^{(L)}} \frac{\partial M_{Lvu}}{\partial \boldtheta} (\hat{\boldz}^{(L-1)}_v, \hat{\boldz}^{(L-1)}_u, \bolde_{vu}, \boldtheta) \| \ge O(\varepsilon)] \le O(\delta), \label{eq: gradient_sampling_1}
\end{align}
\begin{align}
    \textnormal{Pr}[ \| &\sum_{u \in \mathcal{N}(v)} \frac{\partial M_{Lvu}}{\partial \boldz^{(L-1)}_v}(\boldz^{(L-1)}_v, \boldz^{(L-1)}_u, \bolde_{vu}, \boldtheta) \frac{\partial \boldz^{(L-1)}_v}{\partial \boldtheta} \notag \\ &- \frac{\text{deg}(v)}{r^{(L)}} \sum_{u \in \mathcal{S}^{(L)}} \frac{\partial M_{Lvu}}{\partial \hat{\boldz}^{(L-1)}_v}(\hat{\boldz}^{(L-1)}_v, \hat{\boldz}^{(L-1)}_u, \bolde_{vu}, \boldtheta) \frac{\partial \hat{\boldz}^{(L-1)}_v}{\partial \boldtheta} \| \ge O(\varepsilon)] \le O(\delta), \label{eq: gradient_sampling_2}
\end{align}
\begin{align}
    \textnormal{Pr}[ \| &\sum_{u \in \mathcal{N}(v)} \frac{\partial M_{Lvu}}{\partial \boldz^{(L-1)}_u}(\boldz^{(L-1)}_v, \boldz^{(L-1)}_u, \bolde_{vu}, \boldtheta) \frac{\partial \boldz^{(L-1)}_u}{\partial \boldtheta} \notag \\ &- \frac{\text{deg}(v)}{r^{(L)}} \sum_{u \in \mathcal{S}^{(L)}} \frac{\partial M_{Lvu}}{\partial \hat{\boldz}^{(L-1)}_u}(\hat{\boldz}^{(L-1)}_v, \hat{\boldz}^{(L-1)}_u, \bolde_{vu}, \boldtheta) \frac{\partial \hat{\boldz}^{(L-1)}_u}{\partial \boldtheta} \| \ge O(\varepsilon)] \le O(\delta), \label{eq: gradient_sampling_3}
\end{align}
holds true by equations (\ref{eq: grad_Ubound_induction}), (\ref{eq: grad_Mbound_induction}), (\ref{eq: grad_z_induction}), (\ref{eq: grad_h_induction}), (\ref{eq: gradient_hoeffding_1}), (\ref{eq: gradient_hoeffding_2}), and (\ref{eq: gradient_hoeffding_3}).
Therefore, if we take $r^{(1)}, \dots r^{(L)}$ sufficiently large, $\textnormal{Pr}[ \| \frac{\partial \boldz^{(L)}_v}{\partial \boldtheta} - \frac{\partial \boldz^{(L)}_v}{\partial \boldtheta} \|_F \ge \varepsilon] \le \delta$ holds true by equations (\ref{eq: grad_Ubound_induction}), (\ref{eq: grad_Mbound_induction}), (\ref{eq: grad_z_induction}), (\ref{eq: grad_h_induction}), (\ref{eq: gradient_sampling_1}), (\ref{eq: gradient_sampling_2}), and (\ref{eq: gradient_sampling_3}) because the universal constants can be reduced arbitrarily if we increase the constant of the number of samples.

\end{prooftn}

\begin{prooftn}{\ref{gtime}}
We prove this by performing mathematical induction on the number of layers.

\noindent \textbf{Base case:} We show that the statement holds true for $L = 1$.
If $D U_L$ is $K'$-Lipschitz continuous, $\varepsilon' = O(\varepsilon)$ in equation (\ref{eq: grad_Ubound_base}). Therefore, $r^{(L)} = O(\frac{1}{\varepsilon^{2}} \log \frac{1}{\delta})$ is sufficient.

\noindent \textbf{Inductive step:} We show that the statement holds true for $L = l+1$ if it holds true for $L = l$.
If $D U_L$ and $\text{deg}(v) D U_{Lvu}$ is $K'$-Lipschitz continuous, $\varepsilon' = O(\varepsilon)$ in equation (\ref{eq: grad_Ubound_induction}) and $\varepsilon'' = O(\varepsilon)$ in equation (\ref{eq: grad_Mbound_induction}). Therefore, $r^{(L)} = O(\frac{1}{\varepsilon^2} \log \frac{1}{\delta})$ and $r^{(1)}, \dots, r^{(L-1)} = O(\frac{1}{\varepsilon^2} (\log \frac{1}{\varepsilon} + \log \frac{1}{\delta}))$ are sufficient.
\end{prooftn}

\begin{lemma} \label{GATbound}
If Assumptions 1 hold true and $\sigma$ is Lipschitz continuous, $\| \boldz^{(l)}_v \|_2$ and $\| \boldz^{(l)}_v \|_2$ $(l = 1, \dots, L)$ of the GAT model are bounded by a constant
\end{lemma}

\begin{proofln}{\ref{GATbound}}
We prove this by performing mathematical induction on the number of layers. The norm of the input of the first layer is bounded by Assumption 1. If $\| \boldz^{(l-1)}_u \|_2$ and $\| \hat{\boldz}^{(l-1)}_u \|_2$ are bounded for all $u \in \mathcal{V}$, $\| \boldh^{(l)}_v \|_2$ and $\| \hat{\boldh}^{(l)}_v \|_2$ are bounded because $\boldh^{(l)}_v$ and $\hat{\boldh}^{(l)}_v$ are the weighted sum of $\boldz^{(l-1)}_u$ and $\hat{\boldz}^{(l-1)}_u$. Therefore, $\| \boldz^{(l)}_u \|_2$ and $\| \hat{\boldz}^{(l)}_u \|_2$ are bounded because $U_l$ is continuous.
\end{proofln}

\begin{proofpn}{\ref{prop: GAT}}
We prove the theorem by performing mathematical induction on the number of layers $L$.

\noindent \textbf{Base case:} We show that the statement holds true for $L = 1$.
Because $U_L$ is Lipschitz continuous,
\begin{align}
\forall \boldz_v, \boldh_v, \boldh'_v, \boldtheta, \| \boldh_v - \boldh'_v \|_2 < O(\varepsilon) \Rightarrow \|U_{L}(\boldz_v, \boldh_v, \boldtheta) - U_{L}(\boldz_v, \boldh'_v, \boldtheta)\|_2 < \varepsilon \label{eq: GAT_Ubound_base}.
\end{align}
Let $e_u = \exp (\textsc{LeakyReLU}(\bolda^{(l) \top} [\boldW^{(0)} \boldz^{(0)}_v, \boldW^{(0)} \boldz^{(0)}_uu]))$. Then,
\begin{align*}
& \hat{\boldh}_v^{(L)} - \boldh_v^{(L)} \\
&= \sum_{u \in \mathcal{S}^{(L)}} \hat{\alpha}_{vu} \boldz^{(0)}_u - \sum_{u \in \mathcal{N}(v)} \alpha_{vu} \boldz^{(0)}_u  \\
&= \frac{1}{r^{(L)}} \sum_{u \in \mathcal{S}^{(L)}} \frac{e_u}{\frac{1}{r^{(L)}} \sum_{u' \in \mathcal{S}^{(L)}} e_{u'}} \boldz^{(0)}_u - \frac{1}{\text{deg}(v)} \sum_{u \in \mathcal{N}(v)} \frac{e_u}{\frac{1}{\text{deg}(v)} \sum_{u' \in \mathcal{N}(v)} e_{u'}} \boldz^{(0)}_u.
\end{align*}
Let $x_k$ be the $k$-th sample in $\mathcal{S}^{(L)}$ and $X_k = e_{x_k}$. Then, 
\begin{align}
\mathbb{E} [X_k] = \frac{1}{\mathcal{N}(v)} \sum_{u \in \mathcal{N}(v)} e_u \label{eq: GAT_expectation_base}.
\end{align}
There exists a constant $c > 0, C > 0$ such that for any input satisfying Assumption 1, 
\begin{align}
c < | X_k | < C \label{eq: GAT_bound_base},
\end{align}
because $\| \boldz^{(0)}_v \|_2, \| \boldz^{(0)}_{x_k} \|_2, \| \boldW^{(0)} \|_F$, and $\| \bolda^{(0)} \|_2$ are bounded by Assumption 1. 
Therefore, if we take $r^{(L)} = O(\frac{1}{\varepsilon^{2}} \log \frac{1}{\delta})$ samples,
\begin{align*}
\textnormal{Pr}[| \frac{1}{r^{(L)}} \sum_{u \in \mathcal{S}^{(L)}} e_u - \frac{1}{\mathcal{N}(v)} \sum_{u \in \mathcal{N}(v)} e_u | \ge O(\varepsilon)] \le O(\delta)
\end{align*}
by the Hoeffding's inequality and equations (\ref{eq: GAT_expectation_base}) and (\ref{eq: GAT_bound_base}). 
Because $f(x) = 1 / x$ is Lipschitz continuous in $x > c > 0$,
\begin{align}
\textnormal{Pr}[| \frac{1}{\frac{1}{r^{(L)}} \sum_{u \in \mathcal{S}^{(L)}} e_u} - \frac{1}{\frac{1}{\mathcal{N}(v)} \sum_{u \in \mathcal{N}(v)} e_u} | \ge O(\varepsilon)] \le O(\delta) \label{eq: GAT_frac}
\end{align}
Let
\[ Y_k = \frac{e_{x_k}}{\frac{1}{\mathcal{N}(v)} \sum_{u' \in \mathcal{N}(v)} e_{u'}} \boldz^{(0)}_{x_k}. \]
Then, 
\begin{align}
\mathbb{E} [Y_k] = \frac{1}{\mathcal{N}(v)} \sum_{u \in \mathcal{N}(v)} \frac{e_u}{\frac{1}{\mathcal{N}(v)} \sum_{u' \in \mathcal{N}(v)} e_{u'}} \boldz^{(0)}_u \label{eq: GAT_expectation_y_base}.
\end{align}
There exists a constant $C' \in \mathbb{R}$ such that for any input satisfying Assumption 1, 
\begin{align}
\| Y_k \|_2 < C' \label{eq: GAT_bound_y_base}
\end{align}
holds true because $\| \boldz^{(0)}_u \|_2$ are bounded, and $c < | e_u | < C$.
Therefore, if we take $r^{(L)} = O(\frac{1}{\varepsilon^{2}} \log \frac{1}{\delta})$ samples,
\begin{align}
\textnormal{Pr}[\| &\frac{1}{r^{(L)}} \sum_{u \in \mathcal{S}^{(L)}} \frac{e_u}{\frac{1}{\text{deg}(v)} \sum_{u' \in \mathcal{N}(v)} e_{u'}} \boldz^{(0)}_u \notag \\ &\quad - \frac{1}{\text{deg}(v)} \sum_{u \in \mathcal{N}(v)} \frac{e_u}{\frac{1}{\text{deg}(v)} \sum_{u' \in \mathcal{N}(v)} e_{u'}} \boldz^{(0)}_u \|_2 \ge O(\varepsilon)] \le O(\delta) \label{eq: GAT_outer}
\end{align}
holds true by the Hoeffding's inequality and equations (\ref{eq: GAT_expectation_y_base}) and (\ref{eq: GAT_bound_y_base}). 
Therefore,
\begin{align}
\textnormal{Pr}[\| &\frac{1}{r^{(L)}} \sum_{u \in \mathcal{S}^{(L)}} \frac{e_u}{\frac{1}{r^{(L)}} \sum_{u' \in \mathcal{S}^{(L)}} e_{u'}} \boldz^{(0)}_u \notag \\ &\quad - \frac{1}{\text{deg}(v)} \sum_{u \in \mathcal{N}(v)} \frac{e_u}{\frac{1}{\text{deg}(v)} \sum_{u' \in \mathcal{N}(v)} e_{u'}} \boldz^{(0)}_u \|_2 \ge O(\varepsilon)] \le O(\delta) \label{eq: GAT_z}
\end{align}
holds true by the triangle inequality and equations (\ref{eq: GAT_frac}) and (\ref{eq: GAT_outer}), and $\textnormal{Pr}[\|\hat{\boldz}_v^{(L)} - \boldz_v^{(L)}\|_2 \ge \varepsilon] \le \delta$ holds true by equations (\ref{eq: GAT_Ubound_base}) and (\ref{eq: GAT_z}).

\noindent \textbf{Inductive step:} We show that the statement holds true for $L = l+1$ if it holds true for $L = l$. 
Because $U_L$ is Lipschitz continuous,
\begin{align}
\forall \boldz_v, \boldh_v, \boldh'_v, \boldtheta, \| \boldh_v - \boldh'_v \|_2 < O(\varepsilon) \Rightarrow \|U_{L}(\boldz_v, \boldh_v, \boldtheta) - U_{L}(\boldz_v, \boldh'_v, \boldtheta)\|_2 < \varepsilon \label{eq: GAT_Ubound_induction}
\end{align}
holds true.

\begin{align}
\textnormal{Pr}[\| &\frac{1}{r^{(L)}} \sum_{u \in \mathcal{S}^{(L)}} \frac{e_u}{\frac{1}{r^{(L)}} \sum_{u' \in \mathcal{S}^{(L)}} e_{u'}} \boldz^{(L-1)}_u \notag \\ &\quad - \frac{1}{\text{deg}(v)} \sum_{u \in \mathcal{N}(v)} \frac{e_u}{\frac{1}{\text{deg}(v)} \sum_{u' \in \mathcal{N}(v)} e_{u'}} \boldz^{(L-1)}_u \|_2 \ge O(\varepsilon)] \le O(\delta) \label{eq: GAT_outer_induction}
\end{align}
holds true by the same argument as the base step. If we take $r^{(1)}, \dots, r^{(L-1)} = O(\frac{1}{\varepsilon^2} (\log \frac{1}{\varepsilon} + \log \frac{1}{\delta}))$ samples,

\begin{align}
\textnormal{Pr}[\| \hat{\boldz}^{(L-1)}_u - \boldz^{(L-1)}_u \|_2 \ge O(\varepsilon)] \le O(\varepsilon \delta) \label{eq: GAT_z_indcution}
\end{align}
holds true by the induction hypothesis. Therefore, $\textnormal{Pr}[\|\hat{\boldz}_v^{(L)} - \boldz_v^{(L)}\|_2 \ge \varepsilon] \le \delta$ holds true by equations (\ref{eq: GAT_Ubound_induction}), (\ref{eq: GAT_outer_induction}), and (\ref{eq: GAT_z_indcution}).

\end{proofpn}

\begin{proofpn}{\ref{bound}}
We show that one-layer GraphSAGE-GCN whose activation function is not constant cannot be approximated in constant time if $\| x_v \|_2$ or $\| \theta \|_2$ are not bounded. There exists $a, b \in \mathbb{R}$ such that $\sigma(a) \neq \sigma(0)$ because $\sigma$ is not constant. We consider the following two types of inputs:
\begin{itemize}
    \item $G$ is the clique $K_n$, $\boldW^{(1)} = 1$, and $\boldx_i = a$ for all nodes $i \in \mathcal{V}$.
    \item $G$ is the clique $K_n$, $\boldW^{(1)} = 1$, $\boldx_i = a (i \neq v)$ for some $v \in \mathcal{V}$, and $\boldx_v = n (b - a)$.
\end{itemize}
Then, for the former input, $\boldz_v^{(1)} = \sigma(a)$. For the latter type of inputs, $\boldz_v^{(1)} = \sigma(b)$. Let $\mathcal{A}$ be an arbitrary constant algorithm and $C$ be the number of queries $\mathcal{A}$ makes when we set $\varepsilon = |\sigma(a) - \sigma(b)| / 3$. When $\mathcal{A}$ calculates the embedding of $u \neq v \in \mathcal{V}$, the states of all nodes but $u$ are symmetrical until $\mathcal{A}$ makes a query about that node. Therefore, if $n$ is sufficiently large, $\mathcal{A}$ does not make any query about $v$ with high probability (i.e., at least $(1 - \frac{1}{n - 1})^C$). If $\mathcal{A}$ does not make any query about $v$, the state of $\mathcal{A}$ is the same for both types of inputs. If the approximation error is less than $\varepsilon$ for the first type of inputs, the approximation error is larger than $\varepsilon$ for the second type of inputs by the triangle inequality and vice versa. Therefore, $\mathcal{A}$ fails to approximate the embeddings of either type of inputs with the absolute error of at most $\varepsilon$. As for $\boldtheta$, we set $\boldW^{(1)} = n$ and $\boldx_i = a/n$ and $\boldx_x = b - a$. Then, the same argument follows. 
\end{proofpn}

\begin{proofpn}{\ref{normalization}}
We consider the one-layer GraphSAGE-GCN with ReLU and normalization (i.e., $\sigma(\boldx) = \textsc{ReLU}(\boldx) / \|\textsc{ReLU}(\boldx)\|_2$). We use the following two types of inputs:
\begin{itemize}
    \item $G$ is the clique $K_n$, $\boldW^{(1)}$ is the identity matrix $\boldI_2$, $\boldx_i = (0, 0)^\top (i \neq v)$ for some node $v \in \mathcal{V}$, and $\boldx_v = (1, 0)^\top$.
    \item $G$ is the clique $K_n$, $\boldW^{(1)}$ is the identity matrix $\boldI_2$, $\boldx_i = (0, 0)^\top (i \neq v)$ for some node $v \in \mathcal{V}$, and $\boldx_v = (0, 1)^\top$.
\end{itemize}
Then, for the former type of inputs, $\boldh_i = (1/n, 0)^\top$, $\boldz_i = (1, 0)^\top$, and $\frac{\partial z_{i2}}{\partial W_{21}} = 1$ for all $i \in \mathcal{V}$. For the latter type of inputs, $\boldh_i = (0, 1/n)^\top$, $\boldz_i = (0, 1)^\top$, and $\frac{\partial \boldz_{i2}}{\partial W_{21}} = 0$ for all $i \in \mathcal{V}$.
Let $\mathcal{A}$ be an arbitrary constant algorithm and $C$ be the number of queries $\mathcal{A}$ makes when we set $\varepsilon = 1/3$. When $\mathcal{A}$ calculates the embedding or gradient of $u \neq v \in \mathcal{V}$, the states of all nodes but $u$ are symmetrical until $\mathcal{A}$ makes a query about that node. Therefore, if $n$ is sufficiently large, $\mathcal{A}$ does not make any query about $v$ with high probability (i.e., at least $(1 - \frac{1}{n - 1})^C$). If $\mathcal{A}$ does not make any query about $v$, the state of $\mathcal{A}$ is the same for both types of inputs. If the approximation error is less than $\varepsilon$ for the first type of inputs, the approximation error is larger than $\varepsilon$ for the second type of inputs by the triangle inequality and vice versa. Therefore, $\mathcal{A}$ fails to approximate the embeddings and gradients of either type of inputs with the absolute error of at most $\varepsilon$.
\end{proofpn}

\begin{proofpn}{\ref{relu}}
We consider the one-layer GraphSAGE-GCN with ReLU  (i.e., $\sigma(\boldx) = \textsc{ReLU}(\boldx)$). We use the following two types of inputs:
\begin{itemize}
    \item $G$ is the clique $K_n$, $\boldW^{(1)} = (-1, 1)$, $\boldx_i = (1, 1)^\top (i \neq v)$ for some node $v \in \mathcal{V}$, and $\boldx_v = (1, 2)^\top$.
    \item $G$ is the clique $K_n$, $\boldW^{(1)} = (-1, 1)$, $\boldx_i = (1, 1)^\top (i \neq v)$ for some node $v \in \mathcal{V}$, and $\boldx_v = (1, 0)^\top$.
\end{itemize}
Then, for the former type of inputs, $\textsc{MEAN}(\{\boldx_u \mid u \in \mathcal{N}(v) \}) = (1, 1 + \frac{1}{n})^\top$, $\boldh_v = \boldz_v = \frac{1}{n}$, and $\frac{\partial \boldz_v}{\partial \boldW} = (1, 1 + \frac{1}{n})$ for all $i \in \mathcal{V}$. For the latter type of inputs, $\textsc{MEAN}(\{\boldx_u \mid u \in \mathcal{N}(v) \}) = (1, 1 - \frac{1}{n})^\top$, $\boldh_v = -\frac{1}{n}$, $\boldz_v = 0$, and $\frac{\partial \boldz_v}{\partial \boldW} = (0, 0)$ for all $i \in \mathcal{V}$.
Let $\mathcal{A}$ be an arbitrary constant algorithm and $C$ be the number of queries $\mathcal{A}$ makes when we set $\varepsilon = 1/3$. When $\mathcal{A}$ calculates the gradient of $u \neq v \in \mathcal{V}$, the states of all nodes but $u$ are symmetrical until $\mathcal{A}$ makes a query about that node. Therefore, if $n$ is sufficiently large, $\mathcal{A}$ does not make any query about $v$ with high probability (i.e., at least $(1 - \frac{1}{n - 1})^C$). If $\mathcal{A}$ does not make any query about $v$, the state of $\mathcal{A}$ is the same for both types of inputs. If the approximation error is less than $\varepsilon$ for the first type of inputs, the approximation error is larger than $\varepsilon$ for the second type of inputs by the triangle inequality and vice versa. Therefore, $\mathcal{A}$ fails to approximate the gradients of either type of inputs with the absolute error of at most $\varepsilon$.
\end{proofpn}

\begin{proofpn}{\ref{pool}}
We consider the one-layer GraphSAGE-pool whose activation function satisfies $\sigma(1) \neq \sigma(0)$ and the following two types of inputs:
\begin{itemize}
    \item $G$ is the clique $K_n$, $\boldW^{(1)} = 1$, $\boldb = 0$, and $\boldx_i = 0$ for all nodes $v \in \mathcal{V}$.
    \item $G$ is the clique $K_n$, $\boldW^{(1)} = 1$, $\boldb = 0$, $\boldx_i = 0~(i \neq v)$ for some node $v \in \mathcal{V}$, and $\boldx_v = 1$.
\end{itemize}
Then, for the former type of inputs, $\boldz_i = \sigma(0)$ for all $i \in \mathcal{V}$. For the latter type of inputs, $\boldz_i = \sigma(1)$ for all $i \in \mathcal{V}$. 
Let $\mathcal{A}$ be an arbitrary constant algorithm and $C$ be the number of queries $\mathcal{A}$ makes when we set $\varepsilon = |\sigma(1) - \sigma(0)|/3$. When $\mathcal{A}$ calculates the embedding of $u \neq v \in \mathcal{V}$, the states of all nodes but $u$ are symmetrical until $\mathcal{A}$ makes a query about that node. Therefore, if $n$ is sufficiently large, $\mathcal{A}$ does not make any query about $v$ with high probability (i.e., at least $(1 - \frac{1}{n - 1})^C$). If $\mathcal{A}$ does not make any query about $v$, the state of $\mathcal{A}$ is the same for both types of inputs. If the approximation error is less than $\varepsilon$ for the first type of inputs, the approximation error is larger than $\varepsilon$ for the second type of inputs by the triangle inequality and vice versa. Therefore, $\mathcal{A}$ fails to approximate the embeddings of either type of inputs with the absolute error of at most $\varepsilon$.
\end{proofpn}

\begin{proofpn}{\ref{gcn}}
We consider the one-layer GCNs whose activation function satisfies $\sigma(1) \neq \sigma(0)$. We use the following two types of inputs:
\begin{itemize}
    \item $G$ is a star graph, where $v \in \mathcal{V}$ is the center of $G$, $\boldW^{(1)} = 1$, and all features are $0$.
    \item $G$ is a star graph, where $v \in \mathcal{V}$ is the center of $G$, $\boldW^{(1)} = 1$, and the features of $\sqrt{2n}$ leafs are $1$ and the features of other nodes are $0$.
\end{itemize}
Then, for the former type of inputs, $\boldz_v = \sigma(0)$. For the latter type of inputs, $\boldz_v = \sigma(1)$. 
Let $\mathcal{A}$ be an arbitrary constant algorithm and $C$ be the number of queries $\mathcal{A}$ makes when we set $\varepsilon = |\sigma(1) - \sigma(0)|/3$. When $\mathcal{A}$ calculates the embedding of $u \in \mathcal{V}$ that $\boldx_u = 0$, the states of all nodes but $u$ are symmetrical until $\mathcal{A}$ makes a query about that node. Therefore, if $n$ is sufficiently large, $\mathcal{A}$ does not make any query about $v$ with high probability (i.e., at least $(1 - \frac{\sqrt{2n}}{n - 1})^C$). If $\mathcal{A}$ does not make any query about $v$, the state of $\mathcal{A}$ is the same for both types of inputs. If the approximation error is less than $\varepsilon$ for the first type of inputs, the approximation error is larger than $\varepsilon$ for the second type of inputs by the triangle inequality and vice versa. Therefore, $\mathcal{A}$ fails to approximate the embeddings of either type of inputs with the absolute error of at most $\varepsilon$.
\end{proofpn}

\begin{lemma} \label{GCNbound}
If Assumptions 1 and 6 hold true and $\sigma$ is Lipschitz continuous, $\| \boldz^{(l)}_v \|_2$ and $\| \boldz^{(l)}_v \|_2$ $(l = 1, \dots, L)$ of the GCN model are bounded by a constant.
\end{lemma}

\begin{proofln}{\ref{GCNbound}}
We prove this by performing mathematical induction on the number of layers. The norm of the input of the first layer is bounded by Assumption 1. If $\| \boldz^{(l-1)}_u \|_2$ and $\| \hat{\boldz}^{(l-1)}_u \|_2$ are bounded by $B$ for all $u \in \mathcal{V}$, $\| \boldh^{(l)}_v \|_2$ and $\| \hat{\boldh}^{(l)}_v \|_2$ are bounded because under Assumption 6,
\begin{align*}
    \|\boldh^{(l)}_v\|_2 &= \|\sum_{u \in \mathcal{N}(v)} \frac{\boldz^{(l-1)}_u}{\sqrt{\text{deg}(v) \text{deg}(u)}}\| \\
    &\le \frac{1}{\text{deg}(v)} \sum_{u \in \mathcal{N}(v)} \sqrt{\frac{\text{deg}(v)}{\text{deg}(u)}} \|\boldz^{(l-1)}_u\|_2 \\
    &\le \frac{\sqrt{C}}{\text{deg}(v)} \sum_{u \in \mathcal{N}(v)} \|\boldz^{(l-1)}_u\|_2 \\
    &\le B \sqrt{C}.
\end{align*}
\begin{align*}
    \|\hat{\boldh}^{(l)}_v\|_2 &= \frac{\text{deg}(v)}{|\mathcal{S}^{(l)}|} \sum_{u \in \mathcal{S}^{(l)}} \frac{\hat{\boldz}^{(l-1)}_u}{\sqrt{\text{deg}(v) \text{deg}(u)}} \\
    &\le \frac{1}{|\mathcal{S}^{(l)}|} \sum_{u \in \mathcal{S}^{(l)}} \sqrt{\frac{\text{deg}(v)}{\text{deg}(u)}} \|\hat{\boldz}^{(l-1)}_u\|_2 \\
    &\le \frac{\sqrt{C}}{|\mathcal{S}^{(l)}|} \sum_{u \in \mathcal{S}^{(l)}} \|\hat{\boldz}^{(l-1)}_u\|_2 \\
    &\le B \sqrt{C}.
\end{align*}
Therefore, $\| \boldz^{(l)}_u \|_2$ and $\| \hat{\boldz}^{(l)}_u \|_2$ are bounded because $U_l$ is continuous.
\end{proofln}

\begin{proofpn}{\ref{prop: GCN}}
We prove the theorem by performing mathematical induction on the number of layers $L$.

\noindent \textbf{Base case:} We show that the statement holds true for $L = 1$.
Because $U_L$ is Lipschitz continuous,
\begin{align}
\forall \boldz_v, \boldh_v, \boldh'_v, \boldtheta, \| \boldh_v - \boldh'_v \|_2 < O(\varepsilon) \Rightarrow \|U_{L}(\boldz_v, \boldh_v, \boldtheta) - U_{L}(\boldz_v, \boldh'_v, \boldtheta)\|_2 < \varepsilon \label{eq: GCN_Ubound_base}.
\end{align}
Let $x_k$ be the $k$-th sample in $\mathcal{S}^{(L)}$ and $X_k = \sqrt{\frac{\text{deg}(v)}{\text{deg}(x_k)}} \boldz^{(0)}_{x_k}$. Then, 
\begin{align}
\mathbb{E} [X_k] = \frac{1}{\text{deg}(v)} \sum_{u \in \mathcal{N}(v)} \sqrt{\frac{\text{deg}(v)}{\text{deg}(u)}} \boldz^{(0)}_u = \sum_{u \in \mathcal{N}(v)} \frac{\boldz^{(0)}_u}{\sqrt{\text{deg}(v) \text{deg}(u)}} = \boldh^{(L)}_v \label{eq: GCN_expectation_base}.
\end{align}
There exists a constant $C > 0$ such that for any input satisfying Assumption 1, 
\begin{align}
\| X_k \|_2 < C \label{eq: GCN_bound_base},
\end{align}
because $\| \boldz^{(0)}_{x_k} \|_2$ and $\frac{\text{deg}(v)}{\text{deg}(x_k)}$ are bounded by Assumptions 1 and 6. 
Therefore, if we take $r^{(L)} = O(\frac{1}{\varepsilon^{2}} \log \frac{1}{\delta})$ samples, $\textnormal{Pr}[\|\hat{\boldz}_v^{(L)} - \boldz_v^{(L)}\|_2 \ge \varepsilon] \le \delta$ holds by the Hoeffding's inequality and equations (\ref{eq: GCN_Ubound_base}), (\ref{eq: GCN_expectation_base}), and (\ref{eq: GCN_bound_base}). 

\noindent \textbf{Inductive step:} We show that the statement holds true for $L = l+1$ if it holds true for $L = l$. 
Because $U_L$ is Lipschitz continuous and does not use $\boldz_v$,
\begin{align}
\forall \boldz_v, \boldz'_v, \boldh_v, \boldh'_v, \boldtheta, \| \boldh_v - \boldh'_v \|_2 < O(\varepsilon) \Rightarrow \|U_{L}(\boldz_v, \boldh_v, \boldtheta) - U_{L}(\boldz'_v, \boldh'_v, \boldtheta)\|_2 < \varepsilon \label{eq: GCN_Ubound_induction}
\end{align}
holds true.
If we take $r^{(L)} = O(\frac{1}{\varepsilon^{2}} \log \frac{1}{\delta})$ samples,
\begin{align}
\textnormal{Pr}[\|\frac{1}{\text{deg}(v)} \sum_{u \in \mathcal{N}(v)} \sqrt{\frac{\text{deg}(v)}{\text{deg}(u)}} \boldz^{(0)}_u - \frac{1}{|\mathcal{S}^{(L)}|} \sum_{u \in \mathcal{S}^{(L)}} \sqrt{\frac{\text{deg}(v)}{\text{deg}(u)}} \boldz^{(L-1)}_u\| \ge O(\varepsilon)] \le O(\delta) \label{eq: GCN_outer_induction}
\end{align}
holds true by the same argument as the base case. If we take $r^{(1)}, \dots, r^{(L-1)} = O(\frac{1}{\varepsilon^2} (\log \frac{1}{\varepsilon} + \log \frac{1}{\delta}))$ samples,
\begin{align}
\textnormal{Pr}[\| \hat{\boldz}^{(L-1)}_u - \boldz^{(L-1)}_u \|_2 \ge O(\varepsilon)] \le O(\varepsilon \delta) \label{eq: GCN_z_indcution}
\end{align}
holds true by the induction hypothesis. Therefore, $\textnormal{Pr}[\|\hat{\boldz}_v^{(L)} - \boldz_v^{(L)}\|_2 \ge \varepsilon] \le \delta$  holds true by equations (\ref{eq: GCN_Ubound_induction}), (\ref{eq: GCN_outer_induction}), and (\ref{eq: GCN_z_indcution}).

\end{proofpn}

\begin{acks}
This work was supported by JSPS KAKENHI GrantNumber 20H04244 and 21J22490, and the JST PRESTO program JPMJPR165A.
\end{acks}

\bibliographystyle{ACM-Reference-Format}
\bibliography{sample-base}


\begin{thebibliography}{38}

%%% the Web address.

\ifx \showCODEN    \undefined \def \showCODEN     #1{\unskip}     \fi
\ifx \showDOI      \undefined \def \showDOI       #1{#1}\fi
\ifx \showISBNx    \undefined \def \showISBNx     #1{\unskip}     \fi
\ifx \showISBNxiii \undefined \def \showISBNxiii  #1{\unskip}     \fi
\ifx \showISSN     \undefined \def \showISSN      #1{\unskip}     \fi
\ifx \showLCCN     \undefined \def \showLCCN      #1{\unskip}     \fi
\ifx \shownote     \undefined \def \shownote      #1{#1}          \fi
\ifx \showarticletitle \undefined \def \showarticletitle #1{#1}   \fi
\ifx \showURL      \undefined \def \showURL       {\relax}        \fi
% invisible to TeX
\providecommand\bibfield[2]{#2}
\providecommand\bibinfo[2]{#2}
\providecommand\natexlab[1]{#1}
\providecommand\showeprint[2][]{arXiv:#2}

\bibitem[\protect\citeauthoryear{Barabasi and Albert}{Barabasi and
  Albert}{1999}]%
        {BAmodel}
\bibfield{author}{\bibinfo{person}{Albert-Laszlo Barabasi} {and}
  \bibinfo{person}{Reka Albert}.} \bibinfo{year}{1999}\natexlab{}.
\newblock \showarticletitle{Emergence of Scaling in Random Networks}.
\newblock \bibinfo{journal}{\emph{Science}} \bibinfo{volume}{286},
  \bibinfo{number}{5439} (\bibinfo{year}{1999}), \bibinfo{pages}{509--512}.
\newblock


\bibitem[\protect\citeauthoryear{Baskin, Palyulin, and Zefirov}{Baskin
  et~al\mbox{.}}{1997}]%
        {baskin1997neural}
\bibfield{author}{\bibinfo{person}{Igor~I. Baskin},
  \bibinfo{person}{Vladimir~A. Palyulin}, {and} \bibinfo{person}{Nikolai~S.
  Zefirov}.} \bibinfo{year}{1997}\natexlab{}.
\newblock \showarticletitle{A Neural Device for Searching Direct Correlations
  between Structures and Properties of Chemical Compounds}.
\newblock \bibinfo{journal}{\emph{Journal of Chemical Information and Computer
  Sciences}} \bibinfo{volume}{37}, \bibinfo{number}{4} (\bibinfo{year}{1997}),
  \bibinfo{pages}{715--721}.
\newblock


\bibitem[\protect\citeauthoryear{Bruna, Zaremba, Szlam, and LeCun}{Bruna
  et~al\mbox{.}}{2014}]%
        {bruna2013spectral}
\bibfield{author}{\bibinfo{person}{Joan Bruna}, \bibinfo{person}{Wojciech
  Zaremba}, \bibinfo{person}{Arthur Szlam}, {and} \bibinfo{person}{Yann
  LeCun}.} \bibinfo{year}{2014}\natexlab{}.
\newblock \showarticletitle{Spectral Networks and Locally Connected Networks on
  Graphs}. In \bibinfo{booktitle}{\emph{2nd International Conference on
  Learning Representations, {ICLR}}}.
\newblock


\bibitem[\protect\citeauthoryear{Chazelle, Rubinfeld, and Trevisan}{Chazelle
  et~al\mbox{.}}{2005}]%
        {chazelle2005approximating}
\bibfield{author}{\bibinfo{person}{Bernard Chazelle}, \bibinfo{person}{Ronitt
  Rubinfeld}, {and} \bibinfo{person}{Luca Trevisan}.}
  \bibinfo{year}{2005}\natexlab{}.
\newblock \showarticletitle{Approximating the Minimum Spanning Tree Weight in
  Sublinear Time}.
\newblock \bibinfo{journal}{\emph{{SIAM} J. Comput.}} \bibinfo{volume}{34},
  \bibinfo{number}{6} (\bibinfo{year}{2005}), \bibinfo{pages}{1370--1379}.
\newblock


\bibitem[\protect\citeauthoryear{Chen, Ma, and Xiao}{Chen
  et~al\mbox{.}}{2018a}]%
        {FastGCN}
\bibfield{author}{\bibinfo{person}{Jie Chen}, \bibinfo{person}{Tengfei Ma},
  {and} \bibinfo{person}{Cao Xiao}.} \bibinfo{year}{2018}\natexlab{a}.
\newblock \showarticletitle{Fast{GCN}: Fast Learning with Graph Convolutional
  Networks via Importance Sampling}. In \bibinfo{booktitle}{\emph{Proceedings
  of the Sixth International Conference on Learning Representations, {ICLR}}}.
\newblock


\bibitem[\protect\citeauthoryear{Chen, Zhu, and Song}{Chen
  et~al\mbox{.}}{2018b}]%
        {chen2018stochastic}
\bibfield{author}{\bibinfo{person}{Jianfei Chen}, \bibinfo{person}{Jun Zhu},
  {and} \bibinfo{person}{Le Song}.} \bibinfo{year}{2018}\natexlab{b}.
\newblock \showarticletitle{Stochastic Training of Graph Convolutional Networks
  with Variance Reduction}. In \bibinfo{booktitle}{\emph{Proceedings of the
  35th International Conference on Machine Learning, {ICML}}}.
  \bibinfo{publisher}{{PMLR}}, \bibinfo{pages}{941--949}.
\newblock


\bibitem[\protect\citeauthoryear{Chiang, Liu, Si, Li, Bengio, and Hsieh}{Chiang
  et~al\mbox{.}}{2019}]%
        {ClusterGCN}
\bibfield{author}{\bibinfo{person}{Wei{-}Lin Chiang}, \bibinfo{person}{Xuanqing
  Liu}, \bibinfo{person}{Si Si}, \bibinfo{person}{Yang Li},
  \bibinfo{person}{Samy Bengio}, {and} \bibinfo{person}{Cho{-}Jui Hsieh}.}
  \bibinfo{year}{2019}\natexlab{}.
\newblock \showarticletitle{Cluster-GCN: An Efficient Algorithm for Training
  Deep and Large Graph Convolutional Networks}. In
  \bibinfo{booktitle}{\emph{Proceedings of the 25th {ACM} {SIGKDD}
  International Conference on Knowledge Discovery {\&} Data Mining, {KDD}}}.
  \bibinfo{publisher}{{ACM}}, \bibinfo{address}{New York, NY, USA},
  \bibinfo{pages}{257--266}.
\newblock


\bibitem[\protect\citeauthoryear{Czumaj and Sohler}{Czumaj and Sohler}{2004}]%
        {czumaj2004estimating}
\bibfield{author}{\bibinfo{person}{Artur Czumaj} {and}
  \bibinfo{person}{Christian Sohler}.} \bibinfo{year}{2004}\natexlab{}.
\newblock \showarticletitle{Estimating the weight of metric minimum spanning
  trees in sublinear-time}. In \bibinfo{booktitle}{\emph{Proceedings of the
  36th Annual {ACM} Symposium on Theory of Computing, STOC}}.
  \bibinfo{publisher}{{ACM}}, \bibinfo{address}{New York, NY, USA},
  \bibinfo{pages}{175--183}.
\newblock


\bibitem[\protect\citeauthoryear{Defferrard, Bresson, and
  Vandergheynst}{Defferrard et~al\mbox{.}}{2016}]%
        {ChebyNet}
\bibfield{author}{\bibinfo{person}{Micha{\"{e}}l Defferrard},
  \bibinfo{person}{Xavier Bresson}, {and} \bibinfo{person}{Pierre
  Vandergheynst}.} \bibinfo{year}{2016}\natexlab{}.
\newblock \showarticletitle{Convolutional Neural Networks on Graphs with Fast
  Localized Spectral Filtering}. In \bibinfo{booktitle}{\emph{Advances in
  Neural Information Processing Systems 29, NeurIPS}}.
  \bibinfo{pages}{3837--3845}.
\newblock


\bibitem[\protect\citeauthoryear{Erd\H{o}s and R\'{e}nyi}{Erd\H{o}s and
  R\'{e}nyi}{1959}]%
        {ERmodel}
\bibfield{author}{\bibinfo{person}{Paul Erd\H{o}s} {and}
  \bibinfo{person}{Alfr\'{e}d R\'{e}nyi}.} \bibinfo{year}{1959}\natexlab{}.
\newblock \showarticletitle{On Random Graphs {I}}.
\newblock \bibinfo{journal}{\emph{Publicationes Mathematicae}}
  \bibinfo{volume}{6} (\bibinfo{year}{1959}), \bibinfo{pages}{290--297}.
\newblock


\bibitem[\protect\citeauthoryear{Fan, Ma, Li, He, Zhao, Tang, and Yin}{Fan
  et~al\mbox{.}}{2019}]%
        {fan2019graph}
\bibfield{author}{\bibinfo{person}{Wenqi Fan}, \bibinfo{person}{Yao Ma},
  \bibinfo{person}{Qing Li}, \bibinfo{person}{Yuan He},
  \bibinfo{person}{Yihong~Eric Zhao}, \bibinfo{person}{Jiliang Tang}, {and}
  \bibinfo{person}{Dawei Yin}.} \bibinfo{year}{2019}\natexlab{}.
\newblock \showarticletitle{Graph Neural Networks for Social Recommendation}.
  In \bibinfo{booktitle}{\emph{The World Wide Web Conference, {WWW}}}.
  \bibinfo{publisher}{{ACM}}, \bibinfo{address}{New York, NY, USA},
  \bibinfo{pages}{417--426}.
\newblock


\bibitem[\protect\citeauthoryear{Gilmer, Schoenholz, Riley, Vinyals, and
  Dahl}{Gilmer et~al\mbox{.}}{2017}]%
        {MPNNs}
\bibfield{author}{\bibinfo{person}{Justin Gilmer}, \bibinfo{person}{Samuel~S.
  Schoenholz}, \bibinfo{person}{Patrick~F. Riley}, \bibinfo{person}{Oriol
  Vinyals}, {and} \bibinfo{person}{George~E. Dahl}.}
  \bibinfo{year}{2017}\natexlab{}.
\newblock \showarticletitle{Neural Message Passing for Quantum Chemistry}. In
  \bibinfo{booktitle}{\emph{Proceedings of the 34th International Conference on
  Machine Learning, {ICML}}}. \bibinfo{publisher}{{PMLR}},
  \bibinfo{pages}{1263--1272}.
\newblock


\bibitem[\protect\citeauthoryear{Glorot and Bengio}{Glorot and Bengio}{2010}]%
        {xavier}
\bibfield{author}{\bibinfo{person}{Xavier Glorot} {and} \bibinfo{person}{Yoshua
  Bengio}.} \bibinfo{year}{2010}\natexlab{}.
\newblock \showarticletitle{Understanding the difficulty of training deep
  feedforward neural networks}. In \bibinfo{booktitle}{\emph{Proceedings of the
  Thirteenth International Conference on Artificial Intelligence and
  Statistics, {AISTATS}}}. \bibinfo{publisher}{{PMLR}},
  \bibinfo{pages}{249--256}.
\newblock


\bibitem[\protect\citeauthoryear{Gori, Monfardini, and Scarselli}{Gori
  et~al\mbox{.}}{2005}]%
        {gori}
\bibfield{author}{\bibinfo{person}{Marco Gori}, \bibinfo{person}{Gabriele
  Monfardini}, {and} \bibinfo{person}{Franco Scarselli}.}
  \bibinfo{year}{2005}\natexlab{}.
\newblock \showarticletitle{A new model for learning in graph domains}. In
  \bibinfo{booktitle}{\emph{Proceedings of the International Joint Conference
  on Neural Networks, {IJCNN}}}, Vol.~\bibinfo{volume}{2}.
  \bibinfo{pages}{729--734}.
\newblock


\bibitem[\protect\citeauthoryear{Hamilton, Ying, and Leskovec}{Hamilton
  et~al\mbox{.}}{2017}]%
        {GraphSAGE}
\bibfield{author}{\bibinfo{person}{William~L. Hamilton},
  \bibinfo{person}{Zhitao Ying}, {and} \bibinfo{person}{Jure Leskovec}.}
  \bibinfo{year}{2017}\natexlab{}.
\newblock \showarticletitle{Inductive Representation Learning on Large Graphs}.
  In \bibinfo{booktitle}{\emph{Advances in Neural Information Processing
  Systems 30, NeurIPS}}. \bibinfo{pages}{1025--1035}.
\newblock


\bibitem[\protect\citeauthoryear{Hayashi and Yoshida}{Hayashi and
  Yoshida}{2016}]%
        {hayashi2016minimizing}
\bibfield{author}{\bibinfo{person}{Kohei Hayashi} {and} \bibinfo{person}{Yuichi
  Yoshida}.} \bibinfo{year}{2016}\natexlab{}.
\newblock \showarticletitle{Minimizing Quadratic Functions in Constant Time}.
  In \bibinfo{booktitle}{\emph{Advances in Neural Information Processing
  Systems 29, {NeurIPS}}}. \bibinfo{pages}{2217--2225}.
\newblock


\bibitem[\protect\citeauthoryear{Hayashi and Yoshida}{Hayashi and
  Yoshida}{2017}]%
        {hayashi2017fitting}
\bibfield{author}{\bibinfo{person}{Kohei Hayashi} {and} \bibinfo{person}{Yuichi
  Yoshida}.} \bibinfo{year}{2017}\natexlab{}.
\newblock \showarticletitle{Fitting Low-Rank Tensors in Constant Time}. In
  \bibinfo{booktitle}{\emph{Advances in Neural Information Processing Systems
  30, {NeurIPS}}}. \bibinfo{pages}{2470--2478}.
\newblock


\bibitem[\protect\citeauthoryear{Hoeffding}{Hoeffding}{1963}]%
        {Hoeffding}
\bibfield{author}{\bibinfo{person}{Wassily Hoeffding}.}
  \bibinfo{year}{1963}\natexlab{}.
\newblock \showarticletitle{Probability Inequalities for Sums of Bounded Random
  Variables}.
\newblock \bibinfo{journal}{\emph{J. Amer. Statist. Assoc.}}
  \bibinfo{volume}{58}, \bibinfo{number}{301} (\bibinfo{date}{March}
  \bibinfo{year}{1963}), \bibinfo{pages}{13--30}.
\newblock


\bibitem[\protect\citeauthoryear{Huang, Zhang, Rong, and Huang}{Huang
  et~al\mbox{.}}{2018}]%
        {huang2018adaptive}
\bibfield{author}{\bibinfo{person}{Wen{-}bing Huang}, \bibinfo{person}{Tong
  Zhang}, \bibinfo{person}{Yu Rong}, {and} \bibinfo{person}{Junzhou Huang}.}
  \bibinfo{year}{2018}\natexlab{}.
\newblock \showarticletitle{Adaptive Sampling Towards Fast Graph Representation
  Learning}. In \bibinfo{booktitle}{\emph{Advances in Neural Information
  Processing Systems 31, NeurIPS}}.
\newblock


\bibitem[\protect\citeauthoryear{Indyk}{Indyk}{1999}]%
        {indyk1999sublinear}
\bibfield{author}{\bibinfo{person}{Piotr Indyk}.}
  \bibinfo{year}{1999}\natexlab{}.
\newblock \showarticletitle{A Sublinear Time Approximation Scheme for
  Clustering in Metric Spaces}. In \bibinfo{booktitle}{\emph{Proceedings of the
  40th Annual Symposium on Foundations of Computer Science, {FOCS}}}.
  \bibinfo{publisher}{{IEEE}}, \bibinfo{pages}{154--159}.
\newblock


\bibitem[\protect\citeauthoryear{Kipf and Welling}{Kipf and Welling}{2017}]%
        {GCN}
\bibfield{author}{\bibinfo{person}{Thomas~N. Kipf} {and} \bibinfo{person}{Max
  Welling}.} \bibinfo{year}{2017}\natexlab{}.
\newblock \showarticletitle{Semi-Supervised Classification with Graph
  Convolutional Networks}. In \bibinfo{booktitle}{\emph{Proceedings of the
  Fifth International Conference on Learning Representations, {ICLR}}}.
\newblock


\bibitem[\protect\citeauthoryear{Mishra, Oblinger, and Pitt}{Mishra
  et~al\mbox{.}}{2001}]%
        {mishra2001sublinear}
\bibfield{author}{\bibinfo{person}{Nina Mishra}, \bibinfo{person}{Daniel
  Oblinger}, {and} \bibinfo{person}{Leonard Pitt}.}
  \bibinfo{year}{2001}\natexlab{}.
\newblock \showarticletitle{Sublinear time approximate clustering}. In
  \bibinfo{booktitle}{\emph{Proceedings of the Twelfth Annual Symposium on
  Discrete Algorithms, {SODA}}}. \bibinfo{publisher}{{SIAM}},
  \bibinfo{address}{USA}, \bibinfo{pages}{439--447}.
\newblock


\bibitem[\protect\citeauthoryear{Nguyen and Onak}{Nguyen and Onak}{2008}]%
        {nguyen2008constant}
\bibfield{author}{\bibinfo{person}{Huy~N. Nguyen} {and}
  \bibinfo{person}{Krzysztof Onak}.} \bibinfo{year}{2008}\natexlab{}.
\newblock \showarticletitle{Constant-Time Approximation Algorithms via Local
  Improvements}. In \bibinfo{booktitle}{\emph{Proceedings of the 49th Annual
  Symposium on Foundations of Computer Science, {FOCS}}}.
  \bibinfo{publisher}{{IEEE}}, \bibinfo{pages}{327--336}.
\newblock


\bibitem[\protect\citeauthoryear{Park, Kan, Dong, Zhao, and Faloutsos}{Park
  et~al\mbox{.}}{2019}]%
        {GENI}
\bibfield{author}{\bibinfo{person}{Namyong Park}, \bibinfo{person}{Andrey Kan},
  \bibinfo{person}{Xin~Luna Dong}, \bibinfo{person}{Tong Zhao}, {and}
  \bibinfo{person}{Christos Faloutsos}.} \bibinfo{year}{2019}\natexlab{}.
\newblock \showarticletitle{Estimating Node Importance in Knowledge Graphs
  Using Graph Neural Networks}. In \bibinfo{booktitle}{\emph{Proceedings of the
  25th {ACM} {SIGKDD} International Conference on Knowledge Discovery {\&} Data
  Mining, {KDD}}}. \bibinfo{publisher}{{ACM}}, \bibinfo{address}{New York, NY,
  USA}, \bibinfo{pages}{596--606}.
\newblock


\bibitem[\protect\citeauthoryear{Parnas and Ron}{Parnas and Ron}{2007}]%
        {parnas2007approximating}
\bibfield{author}{\bibinfo{person}{Michal Parnas} {and} \bibinfo{person}{Dana
  Ron}.} \bibinfo{year}{2007}\natexlab{}.
\newblock \showarticletitle{Approximating the minimum vertex cover in sublinear
  time and a connection to distributed algorithms}.
\newblock \bibinfo{journal}{\emph{Theor. Comput. Sci.}} \bibinfo{volume}{381},
  \bibinfo{number}{1-3} (\bibinfo{year}{2007}), \bibinfo{pages}{183--196}.
\newblock


\bibitem[\protect\citeauthoryear{Rubinfeld and Sudan}{Rubinfeld and
  Sudan}{1996}]%
        {rubinfeld1996robust}
\bibfield{author}{\bibinfo{person}{Ronitt Rubinfeld} {and}
  \bibinfo{person}{Madhu Sudan}.} \bibinfo{year}{1996}\natexlab{}.
\newblock \showarticletitle{Robust Characterizations of Polynomials with
  Applications to Program Testing}.
\newblock \bibinfo{journal}{\emph{{SIAM} J. Comput.}} \bibinfo{volume}{25},
  \bibinfo{number}{2} (\bibinfo{year}{1996}), \bibinfo{pages}{252--271}.
\newblock


\bibitem[\protect\citeauthoryear{Scarselli, Gori, Tsoi, Hagenbuchner, and
  Monfardini}{Scarselli et~al\mbox{.}}{2009}]%
        {scarselli}
\bibfield{author}{\bibinfo{person}{Franco Scarselli}, \bibinfo{person}{Marco
  Gori}, \bibinfo{person}{Ah~Chung Tsoi}, \bibinfo{person}{Markus
  Hagenbuchner}, {and} \bibinfo{person}{Gabriele Monfardini}.}
  \bibinfo{year}{2009}\natexlab{}.
\newblock \showarticletitle{The Graph Neural Network Model}.
\newblock \bibinfo{journal}{\emph{{IEEE} Trans. Neural Networks}}
  \bibinfo{volume}{20}, \bibinfo{number}{1} (\bibinfo{year}{2009}),
  \bibinfo{pages}{61--80}.
\newblock


\bibitem[\protect\citeauthoryear{Schlichtkrull, Kipf, Bloem, van~den Berg,
  Titov, and Welling}{Schlichtkrull et~al\mbox{.}}{2018}]%
        {RCGN}
\bibfield{author}{\bibinfo{person}{Michael~Sejr Schlichtkrull},
  \bibinfo{person}{Thomas~N. Kipf}, \bibinfo{person}{Peter Bloem},
  \bibinfo{person}{Rianne van~den Berg}, \bibinfo{person}{Ivan Titov}, {and}
  \bibinfo{person}{Max Welling}.} \bibinfo{year}{2018}\natexlab{}.
\newblock \showarticletitle{Modeling Relational Data with Graph Convolutional
  Networks}. In \bibinfo{booktitle}{\emph{The Semantic Web - 15th International
  Conference, {ESWC}}}. \bibinfo{publisher}{Springer},
  \bibinfo{pages}{593--607}.
\newblock


\bibitem[\protect\citeauthoryear{Sperduti and Starita}{Sperduti and
  Starita}{1997}]%
        {sperduti1997supervised}
\bibfield{author}{\bibinfo{person}{Alessandro Sperduti} {and}
  \bibinfo{person}{Antonina Starita}.} \bibinfo{year}{1997}\natexlab{}.
\newblock \showarticletitle{Supervised neural networks for the classification
  of structures}.
\newblock \bibinfo{journal}{\emph{{IEEE} Trans. Neural Networks}}
  \bibinfo{volume}{8}, \bibinfo{number}{3} (\bibinfo{year}{1997}),
  \bibinfo{pages}{714--735}.
\newblock


\bibitem[\protect\citeauthoryear{Vaswani, Shazeer, Parmar, Uszkoreit, Jones,
  Gomez, Kaiser, and Polosukhin}{Vaswani et~al\mbox{.}}{2017}]%
        {vaswani2017attention}
\bibfield{author}{\bibinfo{person}{Ashish Vaswani}, \bibinfo{person}{Noam
  Shazeer}, \bibinfo{person}{Niki Parmar}, \bibinfo{person}{Jakob Uszkoreit},
  \bibinfo{person}{Llion Jones}, \bibinfo{person}{Aidan~N. Gomez},
  \bibinfo{person}{Lukasz Kaiser}, {and} \bibinfo{person}{Illia Polosukhin}.}
  \bibinfo{year}{2017}\natexlab{}.
\newblock \showarticletitle{Attention is All you Need}. In
  \bibinfo{booktitle}{\emph{Advances in Neural Information Processing Systems
  30: Annual Conference on Neural Information Processing Systems 2017,
  {NeurIPS}}}. \bibinfo{pages}{5998--6008}.
\newblock


\bibitem[\protect\citeauthoryear{Veli{\v{c}}kovi{\'c}, Cucurull, Casanova,
  Romero, Li{\`{o}}, and Bengio}{Veli{\v{c}}kovi{\'c} et~al\mbox{.}}{2018}]%
        {GAT}
\bibfield{author}{\bibinfo{person}{Petar Veli{\v{c}}kovi{\'c}},
  \bibinfo{person}{Guillem Cucurull}, \bibinfo{person}{Arantxa Casanova},
  \bibinfo{person}{Adriana Romero}, \bibinfo{person}{Pietro Li{\`{o}}}, {and}
  \bibinfo{person}{Yoshua Bengio}.} \bibinfo{year}{2018}\natexlab{}.
\newblock \showarticletitle{Graph Attention Networks}. In
  \bibinfo{booktitle}{\emph{Proceedings of the Sixth International Conference
  on Learning Representations, {ICLR}}}.
\newblock


\bibitem[\protect\citeauthoryear{Wang, Zhao, Xie, Li, and Guo}{Wang
  et~al\mbox{.}}{2019b}]%
        {KGCN}
\bibfield{author}{\bibinfo{person}{Hongwei Wang}, \bibinfo{person}{Miao Zhao},
  \bibinfo{person}{Xing Xie}, \bibinfo{person}{Wenjie Li}, {and}
  \bibinfo{person}{Minyi Guo}.} \bibinfo{year}{2019}\natexlab{b}.
\newblock \showarticletitle{Knowledge Graph Convolutional Networks for
  Recommender Systems}. In \bibinfo{booktitle}{\emph{The World Wide Web
  Conference, {WWW}}}. \bibinfo{publisher}{{ACM}}, \bibinfo{address}{New York,
  NY, USA}, \bibinfo{pages}{3307--3313}.
\newblock


\bibitem[\protect\citeauthoryear{Wang, He, Cao, Liu, and Chua}{Wang
  et~al\mbox{.}}{2019a}]%
        {KGAT}
\bibfield{author}{\bibinfo{person}{Xiang Wang}, \bibinfo{person}{Xiangnan He},
  \bibinfo{person}{Yixin Cao}, \bibinfo{person}{Meng Liu}, {and}
  \bibinfo{person}{Tat{-}Seng Chua}.} \bibinfo{year}{2019}\natexlab{a}.
\newblock \showarticletitle{{KGAT:} Knowledge Graph Attention Network for
  Recommendation}. In \bibinfo{booktitle}{\emph{Proceedings of the 25th {ACM}
  {SIGKDD} International Conference on Knowledge Discovery {\&} Data Mining,
  {KDD}}}. \bibinfo{publisher}{{ACM}}, \bibinfo{address}{New York, NY, USA},
  \bibinfo{pages}{950--958}.
\newblock


\bibitem[\protect\citeauthoryear{Ying, He, Chen, Eksombatchai, Hamilton, and
  Leskovec}{Ying et~al\mbox{.}}{2018}]%
        {PinSAGE}
\bibfield{author}{\bibinfo{person}{Rex Ying}, \bibinfo{person}{Ruining He},
  \bibinfo{person}{Kaifeng Chen}, \bibinfo{person}{Pong Eksombatchai},
  \bibinfo{person}{William~L. Hamilton}, {and} \bibinfo{person}{Jure
  Leskovec}.} \bibinfo{year}{2018}\natexlab{}.
\newblock \showarticletitle{Graph Convolutional Neural Networks for Web-Scale
  Recommender Systems}. In \bibinfo{booktitle}{\emph{Proceedings of the 24th
  {ACM} {SIGKDD} International Conference on Knowledge Discovery {\&} Data
  Mining, {KDD}}}. \bibinfo{publisher}{{ACM}}, \bibinfo{address}{New York, NY,
  USA}, \bibinfo{pages}{974--983}.
\newblock


\bibitem[\protect\citeauthoryear{Yoshida, Yamamoto, and Ito}{Yoshida
  et~al\mbox{.}}{2009}]%
        {yoshida2009improved}
\bibfield{author}{\bibinfo{person}{Yuichi Yoshida}, \bibinfo{person}{Masaki
  Yamamoto}, {and} \bibinfo{person}{Hiro Ito}.}
  \bibinfo{year}{2009}\natexlab{}.
\newblock \showarticletitle{An improved constant-time approximation algorithm
  for maximum matchings}. In \bibinfo{booktitle}{\emph{Proceedings of the 41st
  Annual {ACM} Symposium on Theory of Computing, {STOC}}}.
  \bibinfo{publisher}{{ACM}}, \bibinfo{address}{New York, NY, USA},
  \bibinfo{pages}{225--234}.
\newblock


\bibitem[\protect\citeauthoryear{Zaheer, Kottur, Ravanbakhsh, P{\'{o}}czos,
  Salakhutdinov, and Smola}{Zaheer et~al\mbox{.}}{2017}]%
        {deepsets}
\bibfield{author}{\bibinfo{person}{Manzil Zaheer}, \bibinfo{person}{Satwik
  Kottur}, \bibinfo{person}{Siamak Ravanbakhsh},
  \bibinfo{person}{Barnab{\'{a}}s P{\'{o}}czos}, \bibinfo{person}{Ruslan
  Salakhutdinov}, {and} \bibinfo{person}{Alexander~J. Smola}.}
  \bibinfo{year}{2017}\natexlab{}.
\newblock \showarticletitle{Deep Sets}. In \bibinfo{booktitle}{\emph{Advances
  in Neural Information Processing Systems 30, NeurIPS}}.
  \bibinfo{pages}{3391--3401}.
\newblock


\bibitem[\protect\citeauthoryear{Zhang, Cui, Neumann, and Chen}{Zhang
  et~al\mbox{.}}{2018}]%
        {DGCNN}
\bibfield{author}{\bibinfo{person}{Muhan Zhang}, \bibinfo{person}{Zhicheng
  Cui}, \bibinfo{person}{Marion Neumann}, {and} \bibinfo{person}{Yixin Chen}.}
  \bibinfo{year}{2018}\natexlab{}.
\newblock \showarticletitle{An End-to-End Deep Learning Architecture for Graph
  Classification}. In \bibinfo{booktitle}{\emph{Proceedings of the
  Thirty-Second {AAAI} Conference on Artificial Intelligence, AAAI}}.
  \bibinfo{publisher}{AAAI Press}, \bibinfo{address}{USA},
  \bibinfo{pages}{4438--4445}.
\newblock


\bibitem[\protect\citeauthoryear{Zou, Hu, Wang, Jiang, Sun, and Gu}{Zou
  et~al\mbox{.}}{2019}]%
        {LADIES}
\bibfield{author}{\bibinfo{person}{Difan Zou}, \bibinfo{person}{Ziniu Hu},
  \bibinfo{person}{Yewen Wang}, \bibinfo{person}{Song Jiang},
  \bibinfo{person}{Yizhou Sun}, {and} \bibinfo{person}{Quanquan Gu}.}
  \bibinfo{year}{2019}\natexlab{}.
\newblock \showarticletitle{Layer-Dependent Importance Sampling for Training
  Deep and Large Graph Convolutional Networks}. In
  \bibinfo{booktitle}{\emph{Advances in Neural Information Processing Systems
  32, NeurIPS}}. \bibinfo{pages}{11247--11256}.
\newblock


\end{thebibliography}

% \appendix

\end{document}